\documentclass[10pt,journal,compsoc]{IEEEtran}
\usepackage{graphicx}
\usepackage{dsfont} 
\usepackage{amsmath}
\usepackage{amssymb}
\usepackage{booktabs}
\usepackage{bbm}
\usepackage{arydshln} 
\usepackage{epsfig}
\usepackage{makecell,rotating}
\usepackage{array}
\usepackage{colortbl}
\usepackage{xcolor}
\usepackage{multirow}
\usepackage{algorithm}  
\usepackage{algpseudocode}

\usepackage[pagebackref,breaklinks,colorlinks]{hyperref}
\usepackage{fancyvrb}
\usepackage{url}

\ifCLASSOPTIONcompsoc

  \usepackage[nocompress]{cite}
\else
  \usepackage{cite}
\fi

\ifCLASSINFOpdf

\else

\fi

\begin{document}

\title{Learning at a Glance: \\Towards Interpretable Data-limited \\Continual Semantic Segmentation via Semantic-Invariance Modelling}

\author{Bo~Yuan,
	Danpei~Zhao*,
	Zhenwei~Shi
\IEEEcompsocitemizethanks{\IEEEcompsocthanksitem Bo Yuan, Danpei Zhao and Zhenwei Shi are with the Image Processing Center, School of Astronautics, Beihang University, Beijing 100191, China. \protect\\
* Corresponding author: Danpei Zhao. \protect\\
E-mail: \{yuanbobuaa, zhaodanpei, shizhenwei\}@buaa.edu.cn. 
}
\thanks{Manuscript received xx, 2023; revised xx.}}

\markboth{Journal of \LaTeX\ Class Files,~Vol.~xx, No.~xx, xx~xx}%
{Shell \MakeLowercase{\textit{et al.}}: Bare Demo of IEEEtran.cls for Computer Society Journals}

\IEEEtitleabstractindextext{%
\begin{abstract}
Continual semantic segmentation (CSS) based on incremental learning (IL) is a great endeavour in developing human-like segmentation models. However, current CSS approaches encounter challenges in the trade-off between preserving old knowledge and learning new ones, where they still need large-scale annotated data for incremental training and lack interpretability. In this paper, we present \textit{Learning at a Glance} (LAG), an efficient, robust, human-like and interpretable approach for CSS. Specifically, LAG is a simple and model-agnostic architecture, yet it achieves competitive CSS efficiency with limited incremental data. Inspired by human-like recognition patterns,  we propose a semantic-invariance modelling approach via semantic features decoupling that simultaneously reconciles solid knowledge inheritance and new-term learning. Concretely, the proposed decoupling manner includes two ways, i.e.,  channel-wise decoupling and spatial-level neuron-relevant semantic consistency. Our approach preserves semantic-invariant knowledge as solid prototypes to alleviate catastrophic forgetting, while also constraining sample-specific contents through an asymmetric contrastive learning method to enhance model robustness during IL steps. Experimental results in multiple datasets validate the effectiveness of the proposed method. Furthermore, we introduce a novel CSS protocol that better reflects realistic data-limited CSS settings, and LAG achieves superior performance under multiple data-limited conditions.
\end{abstract}

}

\maketitle

\IEEEdisplaynontitleabstractindextext

\IEEEpeerreviewmaketitle

\IEEEraisesectionheading{\section{Introduction}
\label{Sec-Introduction}}
\IEEEPARstart{C}{ontinual} learning,  also referred to as incremental learning (IL) or life-long learning, is an approach that focuses on acquiring knowledge in a sequential manner. It enables the learner to perform well on both previously learned tasks and new tasks. Semantic segmentation assigns a label to every pixel in the image. Typically, the popular fully-supervised semantic segmentation methods~\cite{long2015fully, yu2021bisenet} require large-scale annotations to support model training. However, these models are typically designed for closed-set scenes, where they can only handle a fixed number of predefined classes, and all the data needs to be presented to the model at once. 

In real-world scenarios, however, data is usually encountered incrementally, meaning that new data becomes available over time or in a sequential manner. Apparently, discarding the obtained models and re-training new ones on new data signifies a waste of time and computing resources. For example, large language models (LLM) usually cost extreme resources for one-time training. And sometimes the old data can not be accessible due to privacy restrictions and storage burdens. Although recent large-model form~\cite{Kirillov2023SegmentA} achieves fair zero-shot learning ability, they often lack of ability to classify targets with semantic understanding like humans. Additionally, simply re-training the model from scratch can lead to an degradation problem, where the model loses its past ability due to parameter update~\cite{LWF}. As a dense prediction task, continual semantic segmentation (CSS) emerges as a promising but challenging task, with relevance to various practical vision computing fields such as open-world visual interpretation, remote-sensing observation, and autonomous driving.

Current CSS encounters three main challenges: 
1) The issue of catastrophic forgetting and semantic drift, which leads to model degradation over time as new data is introduced.
2) The reliance on large-scale annotated data for incremental learning, which can be time-consuming and costly to obtain.
3) The lack of interpretability in the model updating process, making it difficult to understand why the model fails or being effective and how to improve the IL efficiency.

\begin{figure}[t]
	\centering
	\includegraphics[scale=0.8]{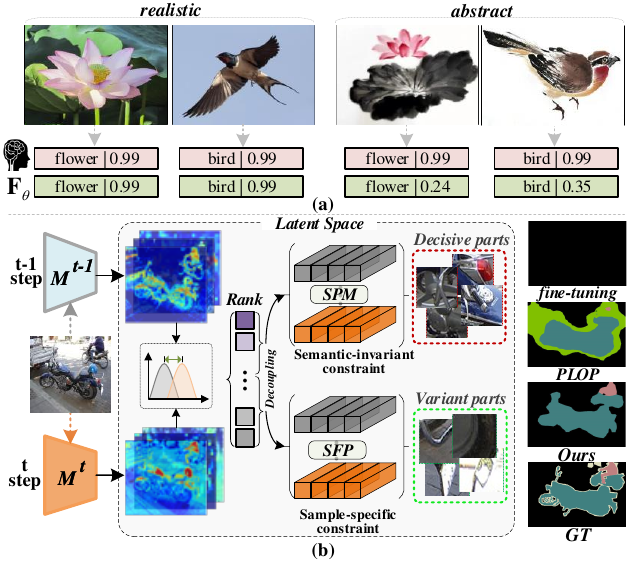}
	\caption{Schematic illustration of the proposed approach. (a) $\textbf{F}_\theta$ is a classifier pretrained on realistic images but without abstract images above. We argue that the model has the potential to make correct classification with key features, which is the inspiration of this paper. (b) The proposed disentangled distillation scheme. Through the disentangled mechanism and the collaboration of corresponding constraints including \textit{SPM} and \textit{SFP}, our model achieves robust CSS performance by alleviating catastrophic forgetting and semantic drift.}
	\label{fig-motivation}
\end{figure}

In this paper, our focus is on developing an efficient, robust, human-like, and interpretable method for CSS tasks. Inspired by the recognition patterns observed in human cognition, we argue that a CSS model should prioritize recognizing and memorizing decisive parts of objects/areas.  As seen in Fig.~\ref{fig-motivation}(a), ordinary individuals are able to recognize both the realistic and abstract representations promptly, demonstrating their ability to process information rapidly and holistically without getting caught up in specific details. However, a classifier $\textbf{F}_{\theta}$ pretrained on realistic images usually struggles when faced with unseen abstract images.

Theoretically, representations with a decomposable structure and consistent semantics associated with different samples/parts within the same class are more likely to generalize to new tasks. However, in current CSS tasks, knowledge transfer often fails since the redundancy and complexity of representations, resulting in model degeneration on both old and new classes. Motivated by human cognition patterns and previous research~\cite{Treisman1980AFT, Judd2009LearningTP}, we propose a novel CSS architecture that incorporates  semantic-invariance modelling (SIM) and solid knowledge transfer methods. 
Fig.~\ref{fig-motivation}(b) depicts the proposed disentangled distillation method. We argue that an object can be classified to decisive parts and variant parts, where the former determine the classification results and the latter contribute to intra-class diversity.  For CSS tasks, we firstly propose two constraints between the old and current semantic representations following semantic-invariance modelling to mutually alleviate the classifier bias caused by catastrophic forgetting and semantic drift. Specifically, we propose a Semantic-invariant Prototype Matching (SPM) module and a Sample-specific Feature Preserving (SFP) module. Functionally, the former focuses on maintaining internal semantic consistency, while the latter addresses intra-class variance to improve model robustness. Additionally,  we propose explicitly modelling an extra \textit{unknown} class to mitigate semantic drift during IL steps. Moreover, we introduce a Neuron-relevant Semantic-consistency Constraint (NSC) based on interpretability analysis. Experimental results on three large-scale datasets prove the effectiveness of the proposed method. Compared with the cumbersome large models, LAG can be trained with lower-budget GPUs. To validate the model's efficiency in realistic applications, we further evaluate its performance with limited incremental training data.

Our contributions are summarized as follows.
\begin{itemize}
\item[$\bullet$] We build an efficient and interpretable model for CSS motivated by human recognition patterns, boosting the model updating efficiency with a semantic-invariance modelling manner. 
\item[$\bullet$] We propose a disentangled distillation method to facilitate knowledge distillation for solid knowledge transfer with interpretability constraints boosting. 
\item[$\bullet$] An explicit unknown class modelling method and a pseudo-labelling manner based on uncertainty are proposed to alleviate background semantic drift and reduce semantic confusion. 
\item[$\bullet$] The proposed model achieves competitive CSS performance in both class incremental and domain incremental scenes across multiple datasets, particularly in practical CSS tasks with limited incremental data.
\end{itemize}

\section{Related Work}
In this section, we review the researches on incremental learning, continual semantic segmentation, disentangled representation learning and explainable AI, respectively. 

\subsection{Incremental Learning}
Incremental learning is a technique that allows a neural network to continuously update its parameters with incremental data, breaking the traditional one-off training process in deep learning. IL has been explored in various fields, including computer vision~\cite{Liu2022IncrementalLW, Belouadah2021ACS}, natural language processing~\cite{Autume2019EpisodicMI, Biesialska2020ContinualLL} and remote sensing~\cite{Bhat2021CILEANETCI}.The most significant challenge in IL is catastrophic forgetting, which occurs when the parameter updates result in the loss of previously learned knowledge. This phenomenon was first identified and discussed as early as the 1980s by McCloskey, et al.~\cite{McCloskey1989CatastrophicII}. That is algorithms trained with backpropagation suffer from severe degeneration just like human suffers from gradual forgetting of previously learned tasks. 
To mitigate this problem, a range of researches~\cite{RainbowMC, IL2MCI, UsingHT, ContinualLO} propose to retrospect known knowledge including sample selection as exemplar memory~\cite{Aljundi2019GradientBS, Han2018CoteachingRT, Rolnick2019ExperienceRF, Fini2020OnlineCL, Shi2021ContinualLV}, prototypes guidance~\cite{Ho2021PrototypesGuidedMR, Ahn2019UncertaintybasedCL, Zhu2021PrototypeAA, Zhu2021SelfPromotedPR}, meta learning~\cite{Javed2019MetaLearningRF, Wang2020EfficientML, Banayeeanzade2021GenerativeVD, Hurtado2021OptimizingRK}, generative adversarial learning~\cite{Ebrahimi2020AdversarialCL, Xiang2019IncrementalLU, Verma2021EfficientFT}, etc. These approaches can achieve effective continual learning performances but normally require extra memory to store old data. However, a more challenging yet practical IL setting arises when no old data is available. In recent years, some researches concentrate on IL without data-replay by using knowledge distillation~\cite{ABD, Kang2022ClassIncrementalLB, Douillard2020PODNetPO, Ye2020DistillingCK, Tian2021RelationshipPreservingKD, Xu2022DeepNN, Zhao2022EmbeddedSI, Hou2018LifelongLV}, conducting weight transfer~\cite{Ahn2019VariationalID, Slim2022DatasetKT} and network architecture extension~\cite{Kanakis2020ReparameterizingCF, Liu2021AdaptiveAN, Yan2021DERDE}, etc.  The primary concern in current IL research is to strike a balance between preventing forgetting and assimilating new knowledge, which is the main focus of this paper.

\subsection{Continual Semantic Segmentation}
Continual semantic segmentation encompasses several types of tasks including class incremental and domain incremental~\cite{yuan2023survey}. CSS encounters challenges such as catastrophic forgetting and semantic drift, which arise from the absence of old data and parameter updates~\cite{Hu2021DistillingCE, Kaushik2021UnderstandingCF, RW}. Based on the usage of exemplar memory, CSS approaches can be categorized into two groups~\cite{yuan2023survey}. The first kind, including methods like~\cite{ILT, CIL, MiB, PLOP, UCD, REMINDER, RCIL}, utilizes knowledge distillation~\cite{Hinton2015DistillingTK, 9115859, Wang2022KnowledgeDA} to inherit the capability of the old model. In the field of remote sensing, research focuses on enhancing small objects~\cite{Li2022ClassIncrementalLN} and multi-level distillation~\cite{Tasar2019IncrementalLF, Shan2022ClassIncrementalLF, qiu2023sats, IDEC}. The second kind involves storing a portion of past training data as exemplar memory. Rebuffi~et al.~\cite{iCaRL} conducts incremental steps with the supervision of representative past training data, but this approach updates the old model, which may exacerbate catastrophic forgetting. Cha~et al.~\cite{SSUL} propose a class-imbalanced sampling strategy and uses a saliency detector to filter out unknown classes. Maracani~et al.~\cite{RECALL} rely on a generative adversarial network or web-crawled data to retrieve images. Recently more efficient sample selection strategies have been proposed~\cite{zhu2023continual, ramasesh2022effect}.  However, these data-replay-based methods impose additional memory consumption, which is critical especially in CSS circumstances. In recent years, few-shot/zero-shot approaches~\cite{Cheraghian2021SemanticawareKD, Xu2020ProgressiveDF} have been explored to reduce data and annotation dependency in IL steps. Whereas these methods typically exhibit low performance in complex visual tasks. In this paper, we propose an exemplar memory-free and data-replay-free approach to address CSS, aiming to meet the requirements of realistic applications.

\subsection{Disentangled Representation Learning}
Disentangled representation learning aims to model the factors of data variations. Bengio~et al.~\cite{Bengio2012RepresentationLA} argue that latent units are sensitive to changes in single generative factors while being relatively invariant to changes in other factors. In computer vision, decoupled learning have been applied in image generation~\cite{Alharbi2020DisentangledIG, Wu2020StyleSpaceAD}, image translation~\cite{Lee2018DRITDI, Liu2021SmoothingTD},  image denoising~\cite{Miao2021HyperspectralDU, Huang2022Neighbor2NeighborAS} and domain adaptation~\cite{BAFFT}, etc. Recent Yeh~et al.~\cite{Yeh2022DecoupledCL} propose a decoupled contrastive loss based on self-supervised augmentation to optimize SimCLR~\cite{SimCLR}. In IL tasks, \cite{9877899} aims to obtain class-disentangled features for knowledge preservation and~\cite{DKD} proposes a decomposed knowledge distillation loss. Some methods use contrastive learning on disentangled latent representations~\cite{SDR,  IDEC}. However, current methods overlook the inner distribution and interpretability of semantic features during IL steps, which we believe is a critical issue to cope with catastrophic forgetting and semantic drift in CSS.

\subsection{Explainable Artificial Intelligence}
Explainable artificial intelligence (XAI) is an emerging field that aims to address the lack of interpretability in traditional machine learning and deep learning algorithms. It has been explored in various fields including image analysis~\cite{Zhang2020ExtractionOA, tjoa2020survey}, object detection~\cite{Zhang2017DeepVotingAR}, autonomous driving~\cite{Abukmeil2021TowardsES}, etc. In computer vision field, the interpretability methods can be mainly divided into three kinds. The first kind is the local/global interpretability methods, which aim to explain the behavior of local features or the entire model. Techniques like LIME~\cite{Ribeiro2016WhySI} construct an approximate linear model to explain the predictions of individual samples.The second kind involves visualization techniques that provide insights into the decision-making process within the model. Representative methods are Grad-CAM~\cite{GradCAM, GradCAM++}, Guided Backpropagation~\cite{Springenberg2014StrivingFS} and LRP~\cite{LRP}, etc.  The third kind includes adversarial methods that aim to explore the boundaries and limitations of the model's decision-making. Techniques like adversarial attacks ~\cite{Slack2019FoolingLA}  perturb the input data to find the minimum changes required to alter the model's output, revealing vulnerabilities and potential biases. However, above approaches focus on interpretation of model outputs, yet leveraging interpretability to enhance the model's ability is still an ongoing challenge.

\section{Methodology}
\label{Sec-Methodology}
\begin{figure*}[t]
	\centering
	\includegraphics[scale=0.94]{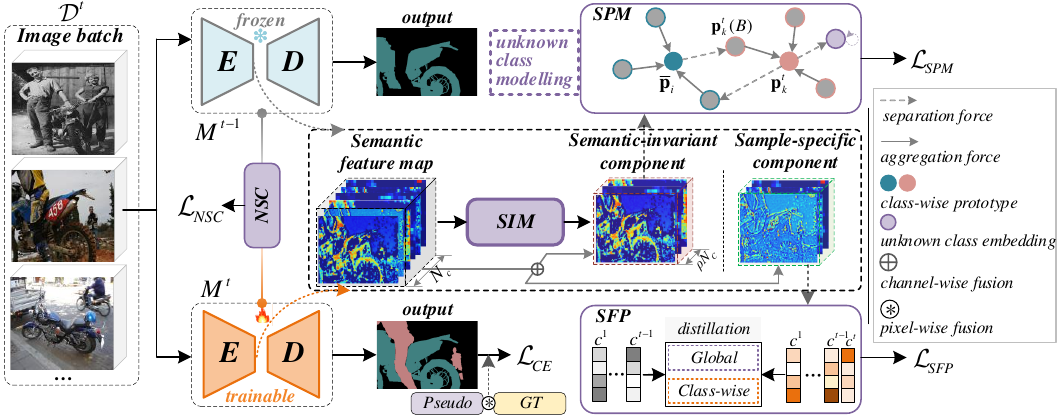}
	\caption{The pipeline of the proposed method at \textit{t} step. An old class (\textit{motorbike}) and a new class (\textit{person}) are used as an example. At each incremental step, the old model is frozen and no old data can be accessed. While the current model is trained only on the incremental data. The neuron-relevant semantic-consistency constraint (NSC) is performed between $M^{t-1}$ and $M^t$. The semantic features are disentangled to semantic-invariant term and sample-specific term in the latent space. Prototype alignment (SPM) and Feature preserving (SFP) modules are designed respectively but trained jointly for disentangled knowledge transfer.}
	\label{fig-network}
\end{figure*}
\subsection{Preliminaries}
Let $\mathcal{D}=\{(x_i, y_i)\}$ signifies the training dataset, where $x_i \in \mathbb{R}^{C\times H\times W}$ denotes the training image and $y_i \in \mathbb{R}^{H\times W}$ denotes the corresponding ground truth. $\mathcal{D}^t$ represents the training dataset for \emph{t} step. At $t$ step, $C^{0:t-1}$ indicates the previously learned classes and $C^t$ indicates the classes for incremental learning. When training on $\mathcal{D}^t$, the training data of old classes, i.e., $\{\mathcal{D}^0, \mathcal{D}^1, \cdots, \mathcal{D}^{t-1}\}$ is inaccessible. And the ground truth in $\mathcal{D}^t$ only covers $C^t$. The complete training process consists of \{Step-0, Step-1, $\cdots$, Step-T\} steps. 
At $t$ step, we use $M^{t-1}$ and $M^t$ to represent the \textit{t-1} and \textit{t} step model, respectively. Let \textbf{F}$_\theta$ indicates the feature extractor.  $N_l$ is the layer number of the feature extractor network.  

\subsection{Semantic Invariance Modelling}
\label{Sec-SFD}
The pipeline of the proposed method is depicted in Fig.~\ref{fig-network}. At each incremental step, since the lack of annotations, all the other classes are actually labelled as \textit{background} (bg) except for the current classes $C^t$. Thus the connotation of \textit{bg} would contain known classes and future classes simultaneously, leading to classifier confusion. To maintain the output consistency from the old and the current models, we propose to disentangle the semantic features in the latent space into semantic-invariant term and sample-specific term, respectively. We argue that the former denotes the semantic content associated with a specific class. While the latter enhances the model generalization. 
Let $\mathcal{F}_c^t \in \mathbb{R}^{H_d\times W_d}$ indicate the feature map produced by $M^t$ at $c$ dimension. $N_c$ represents the channel dimension.
The assumption is that the model tends to make predictions via semantic-invariant features while the sample-specific component contributes to intra-class variance.  We firstly transform $\mathcal{M}_c^t$ to one-dimensional tensors, i.e., $\mathcal{E}_c^{t} \in \mathbb{R}^{H_dW_d}$.

\subsubsection{Channel-wise Decoupling}
Since the old model and current model share the same architecture but with different weights, we take advantage of the channel-wise feature maps to determine the semantic-invariant embeddings from channel level. Specifically, for the embeddings $\{\textbf{e}^{t-1}, \textbf{e}^t\}$ from \{$M^{t-1}, M^t$\},  let $\mathcal{M}_c^t \in \mathbb{R}^{H_d\times W_d}$ indicates the feature map produced by $M^t$ at $c$ dimension. $N_c$ represents the channel dimension. The channel-wise relevance between $M^{t-1}$ and $M^{t}$ can be measured by the similarity:
\begin{equation}
	\mathcal{S}(\mathcal{E}_c^{t}, \mathcal{E}_c^{t-1}) = d(\mathcal{E}_c^{t}, \mathcal{E}_c^{t-1})
	\label{eqn-similarity}
\end{equation}
Specifically, in the image batch, we split the channel-wise embeddings into stable ones and unstable ones by ranking the similarity between the embeddings from $M^{t-1}$ and $M^t$:
\begin{equation}
	\mathcal{R}_{\mathcal{E}} = \{r_1^t, r_2^t, \cdots, r_{N_c}^t \} = \Phi(\mathcal{S},t) 
\end{equation}	
where $r_c^t$ represents the similarity between $\mathcal{E}_c^{t}$ and $\mathcal{E}_c^{t-1}$ in \textit{c}-th dimension at $t$ step. $\Phi$ is the operation to rank the similarity according to Eqn.~\ref{eqn-similarity}.  Let $\mathfrak{E}_{SI}$ and $\mathfrak{E}_{SS}$ denote embeddings assigned to semantic-invariant term and sample-specific term, respectively. In this paper, we assume they meet the additive rule, which means $\textbf{e}^t$: $\textbf{e}^t= \mathfrak{E}_{SI}^t\oplus \mathfrak{E}_{SS}^t$. Specifically, the channel dimension of $ \mathfrak{E}_{SI}^t$ is $\rho N_c$. $\oplus$ indicates the channel-wise concatenation.

\subsubsection{Neuron-relevant Semantic Consistency}
\label{Sec-NSC}
\begin{figure}[tbp]
	\centering
	\includegraphics[scale=0.47]{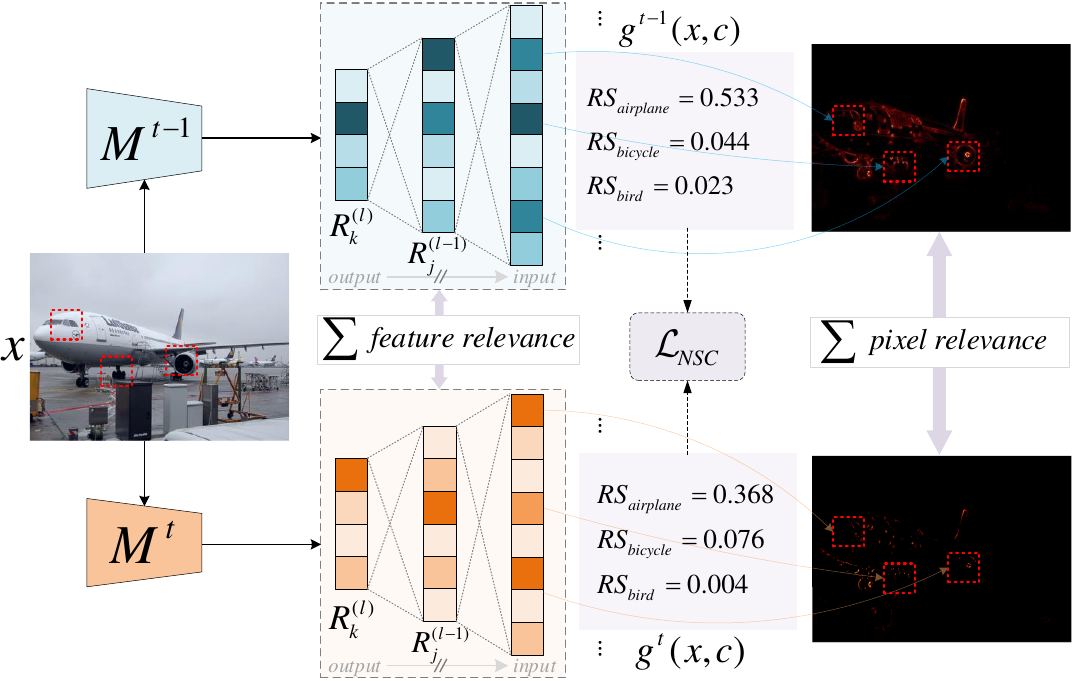}
	\caption{The neuron-relevant semantic-consistency constraint at $t$ step. The feature relevance constraint is the neuron relevance between $M^{t-1}$ and $M^t$. The pixel relevance illustrates the semantic consistency between neurons.}
	\label{fig-NSC}
\end{figure}

To tackle the catastrophic forgetting, we take the benefits of model interpretability to boost the IL efficiency from spatial level. For dense prediction task, we consider the pixel-wise contribution to the model outputs can be revealed by the neuron relevance. 
Our intuitive idea is that model interpretability can be posed as the capability of exploring the latent space of a higher level task (e.g., segmentation) in a principled way, and of capturing the sufficient statistics in the latent space. As seen in Fig.~\ref{fig-NSC}, for each class in $C^{0:t-1}$, we define the spatial-level semantic-invariant features using the results of the layer-wise relevance propagation (LRP)~\cite{LRP}. LRP assigns relevance scores to neurons in consecutive layers of the neural network, indicating their contribution to the final prediction.
Specifically, let $j$ and $k$ represent neurons in two consecutive layers of the neural network. The propagation of relevance score (RS) $R_j^{(l)}$ from layer $l$ to layer $l-1$ is achieved using the following rule:
\begin{equation}
	R_j^{(l-1)} = \sum_k \frac{a_{j,k}^{(l)}}{\sum_j a_{j,k}^{(l)}} R_k^{(l)}
\end{equation}
where $a_{j,k}^{(l)}$ represents the activation between neurons $j$ and $k$ at layer $l$. By iteratively applying this rule from the output layer to the input layer, we can obtain the relevance scores for each pixel, indicating their importance for the final prediction.  Here we propose the neuron relevance between $M^{t}$ and $M^{t-1}$ should be consistent with $C^{0:t-1}$, such the neuron relevance score at $t$ step for class $c$ is obtained by:
\begin{equation}
g(x,c) = \sum_l R^{(l)}  \in \mathbb{R}^{N_c},  c \in C^{0:t-1}
\end{equation}
\begin{algorithm}[htbp]  
	\caption{Pseudocode of neuron-relevant semantic-consistency constraint at $t$ step. } 
	\label{alg-Pseudocode-NSC}  
	\begin{algorithmic}[1] 
		\Require  Training image $x$, $M^{t-1}$, $M^t$
		\Ensure  
		NSC loss $\mathcal{L}_{NSC}$
		\State  Initialize $\mathcal{L}_{NSC} \leftarrow$ 0, relevance score $R \leftarrow$ 0
		\State $ \textbf{e}^{t} \leftarrow M^t(x)$,  $ \textbf{e}^{t-1} \leftarrow M^{t-1}(x)$ \{forward pass\}
		\State \textbf{for} $c \in C^{0:t-1}$ do:
		\State \qquad Calculate $g^{t-1}(x,c), g^{t}(x,c) $ according to Eqn.(4)
		\State \qquad \textbf{for} $l \in N_l$ do:
		\State \qquad \qquad $g^{t-1}(x,c) \leftarrow \sum_k R_{j\leftarrow k}^{(l-1,l)} \leftarrow \textbf{e}^{t-1}$
		\State \qquad \qquad $g^{t}(x,c) \leftarrow   \sum_k  R_{j\leftarrow k}^{(l-1,l)} \leftarrow \textbf{e}^{t}$ 
		\State \qquad \textbf{end  for}
		\State \qquad Calculate NSC loss for class $c$ according to Eqn.(5)
		\State \qquad $d_c \leftarrow ||g^{t-1}(x,c)-g^t(x,c)||_2^2$
		\State \textbf{end for}	
		\State $\mathcal{L}_{NSC} = \sum_c d_c$
		\State Return $\mathcal{L}_{NSC}$
	\end{algorithmic}  
\end{algorithm}   

The prior is that the old model always achieves superior performance to the current model on the old classes due to the catastrophic forgetting during IL steps. For each class in $C^{0:t-1}$, the ideal situation is that the current model achieves the same interpretability, i.e., the semantic-invariant term should be consistent within each learned class. Thus we propose a semantic-consistency constraint (NSC) based on neuron-relevance: 
\begin{equation}
\mathcal{L}_{NSC} =\frac{1}{|C^{0:t-1}|} \sum_c^{C^{0:t-1}} \Vert g^{t-1}(x,c)-g^{t}(x,c) \Vert_2^2
	\label{eqn-NSC}
\end{equation}
where $g^{t-1}(x,c)$ and $g^t(x,c)$ indicates the relevance scoring results from $M^{t-1}$ and $M^t$, respectively.  
The pseudocode of gradient-weighted decoupling manner is shown in Algorithm~\ref{alg-Pseudocode-NSC}.

\subsection{Disentangled Distillation}
\subsubsection{Uncertainty-aware Unknown Class Modelling}
\begin{figure}[t]
	\centering
	\includegraphics[scale=0.5]{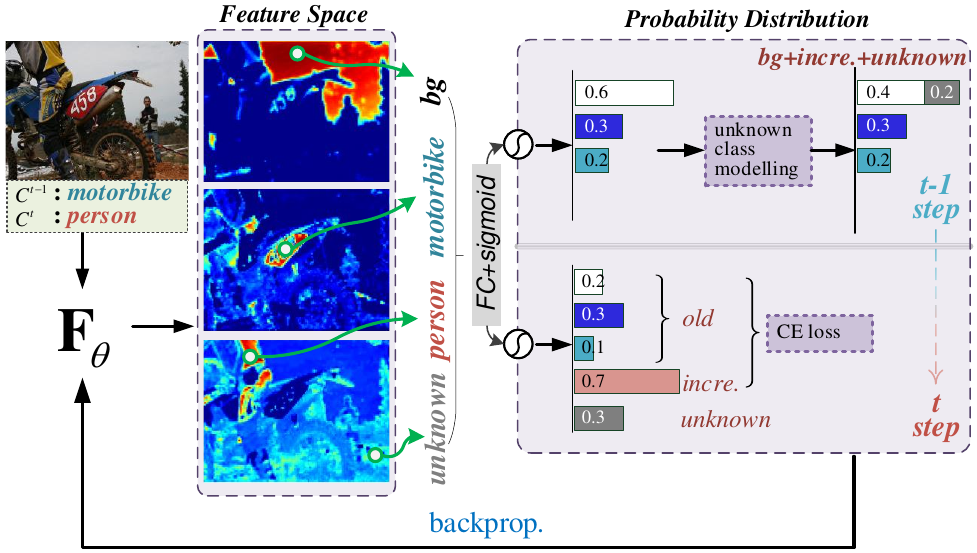}
	\caption{Uncertainty-aware unknown class modelling.}
	\label{fig-unknown_modeling}
\end{figure}

At IL steps, the semantic of \textit{background} (\textit{bg}) class includes the true \textit{bg}, \textit{old}, and \textit{future} classes simultaneously, which results in semantic confusion. To alleviate this problem, we propose an explicit extra \textit{unknown} class.  The pioneering work~\cite{SSUL} models the unknown area via an independent saliency detector as an additional branch, which limits the model conciseness and adaptability to different datasets. Our motivation, as depicted in Fig.~\ref{fig-unknown_modeling}, places emphasis on generality and models the \textit{unknown} class $c_u$ by prediction isolation directing at the true \textit{bg} and without extra branch. Since the semantic of \textit{unknown} class varies at each incremental step, we define probabilities of \textit{unknown} explicitly. At $t$-step training, the prediction of pixel $i$ is obtained by:
\begin{equation}
	\hat{y}_i^t = \mathop{\arg\max}_{c\in \{C^{0:t}, c^b\}} [\textbf{F}_\theta(x_i)]
\end{equation}

The prior of assigning the label as \textit{unknown} at $t$ step is 1) the prediction is classified to $c^b$; 2) even the highest probability $p_i^h$ is lower than the confidence threshold; 3) the top-2 $p_i$ have nearly equal values.
Thus for pixel $i$ in soft segmentation map $\mathcal{M}_s$, 
the prediction within $c^b$ and conform to the above conditions is classified as unknown pixel. At each IL step, the unknown pixel is updated by:
\begin{equation}  
	c^u \leftarrow [\hat{y}_i^t=c^b] \land [p^{h}_i<\Gamma] \land [u_i / \triangle_i <\zeta]
\end{equation}
where $u_i$ and $ \triangle_i$ are the certainty and uncertainty range of pixel $i$, $\Gamma$ and $\zeta$ are used to support above priors as introduced in Sec.~\ref{Sec-UPL}.  
Particularity, the \textit{unknown} classes do not participate in gradient back propagation, as the segmentation loss (e.g., cross-entropy loss) is limited to old and new classes, i.e., $\{C^{0:t-1}\cup C^t\}$. 

\subsubsection{Semantic-invariant Prototype Matching}
\label{Sec-SPM}
Based on the emphasis of the semantic-invariant features, a disentangled distillation manner is  a certain way to resolve the model degradation during IL steps. We design specific supervisions for semantic-invariant and sample-specific features, respectively. For the former, a prototype is maintained for each class belonging to $\{C^{0:t-1} \cup C^t \}$.  To achieve the dismissal of semantic confusion as the IL step increasing, we calculate prototype matching within the same class and differentiation across different classes at each  IL step. Specifically, at \textit{t} step, the prototype for class \textit{k} is defined as $\textbf{p}_k^t$. Comparing $\textbf{p}_k^t$ with the previous prototype $\overline{\textbf{p}}_k$ to calculate the prototype matching loss:
\begin{equation}
	\mathcal{L}_{SPM} = \mathcal{L}_{SPM}^{intra} + \mathcal{L}_{SPM}^{inter}
	\label{eqn-SPM}
\end{equation}
\begin{equation}
	\mathcal{L}_{SPM}^{intra} = \frac{1}{\lvert C^{0:t-1}\rvert} \sum_k^{C^{0:t-1}} d(\textbf{p}_k^t , \overline{\textbf{p}}_k) 
\end{equation}
\begin{equation}
	\mathcal{L}_{SPM}^{inter} = \frac{1}{\lvert C^{0:t}\rvert} \sum_{k,i\neq k}^{ C^{0:t}} \frac{1}{\lVert \hat{\textbf{p}}_k^t-\overline{\textbf{p}}_i \rVert_2^2} + \frac{1}{\lVert\textbf{p}_k^t - \textbf{p}_{bg}\lVert_2^2}
	\label{Eqn-SPM_inter}
\end{equation}
where $\lvert C^{0:t-1}\rvert$ and $\lvert C^{0:t}\rvert$ are the learned class number at \textit{t-1} step and \textit{t} step, respectively. $d(\cdot)$ is a distance measure.  $\textbf{p}_{bg}$ represents the prototype of \textit{background}.  To further facilitate the optimization process, the prototypes are computed in place with a running average updated at each training step using masked average pooling. At \textit{t} step, $\textbf{p}_k^t$ is calculated by:
\begin{equation}
	\textbf{p}_k^t  = \frac{1}{\lvert N^k \rvert}  \sum_{i\in |N^k|} 
								\frac{    \sum_{k} \mathfrak{E}_{{SI}_i^t} \mathbbm{1}[\mathcal{M}_i=k]    }{     \sum_{k} \mathbbm{1}[\mathcal{M}_i=k]       }
\end{equation}
where $\lvert N^k \rvert$ indicates the sample number of class $k$ at each iteration. $\mathbbm{1}$ is an indicator function. $\mathcal{M}_i$ is the mask for sample $i$.  

Specifically, to force the class-wise prototypes distinguished from \textit{bg}, the prototype of \textit{bg} is defined as the average of all class-wise prototypes and unknown semantics.  Thereupon the class-wise prototypes $\textbf{p}_k^t $ is updated following Eqn.~\ref{Eqn-SPM_inter}. However, the known and future classes are mixed in \textit{bg} at the current step due to lack of annotations, which leads to semantic drift. Therefore, $\textbf{p}_{bg}$ is recalculated and the prototypes in $C^{0:t}$ are updated after each incremental training step to reduce semantic drift.

\subsubsection{Sample-specific Feature Preserving}
\label{Sec-SFP}
The SFP module is designed to deal with sample-specific terms to enhance the model robustness and reduce classifier bias on new classes. Let ${\mathfrak{E}_{SS}}_{k}$ represents the variance content within in class \textit{k}. Since the class number may increases at incremental steps, we propose an asymmetric distillation mechanism with a semi-supervised contrastive learning manner. The pioneer~\cite{IDEC} proposes a contrastive learning manner at IL steps but consumes large amounts of time. Our proposed manner is more specific and focuses on the distinguish sample-specific features. Specifically, the distillation consists of global output logits and class-wise semantic embeddings ${\mathfrak{E}_{SS}}_k$, $k\in C^{0:t}$. For the former, the distance between outputs from $M^{t}$ and $M^{t-1}$ is optimized to maintain the model encoding consistency. For the latter, we propose a triplet contrastive distillation manner to force the model to differentiate the embeddings between old and new classes. The optimization goal is defined as: 
\begin{equation}
	\mathcal{L}_{SFP} = d(\textbf{e}^t, \textbf{e}^{t-1}) + D_{CL}\left({\mathfrak{E}_{SS}}_{a}^t, {\mathfrak{E}_{SS}}_{p}^t, {\mathfrak{E}_{SS}}_{n}^t\right) 
	\label{eqn-SFP}
\end{equation}
\begin{equation}
	\begin{aligned}
		&D_{CL}\left({\mathfrak{E}_{SS}}_{a}^{t-1}, {\mathfrak{E}_{SS}}_{p}^t, {\mathfrak{E}_{SS}}_{n}^t\right) = \frac{1}{\lvert C^{0:t} \rvert} \sum_k^{C^{0:t} } max(\\ &d({\mathfrak{E}_{SS}}_{ak}^{t-1}, {\mathfrak{E}_{SS}}_{pk}^t)- d({\mathfrak{E}_{SS}}_{ak}^{t-1}, {\mathfrak{E}_{SS}}_{nk}^t)+ m, 0)
	\end{aligned}
\end{equation}
where ${\mathfrak{E}_{SS}}_{ak}^{t-1}$, ${\mathfrak{E}_{SS}}_{pk}^t$ and ${\mathfrak{E}_{SS}}_{nk}^t$ represent the \textit{anchor}, \textit{positive} and \textit{negative} embeddings belonging to \textit{k}-th class. $m$ is a constant. The asymmetry manifests itself in two ways. One is the inequality of class number at the current step and the previous step. The other is the embedding pairing manner for contrastive learning. Specifically, we select \textit{anchor} embedding from $M^{t-1}$ within old classes to reduce prediction error since catastrophic forgetting. While \textit{positive} and \textit{negative} embeddings are from $M^t$ to optimize the current model. $d(\cdot)$ is a distance measurement function.

The integrated optimization goal of the disentangled distillation is:
\begin{equation}
	\mathcal{L}_{DD} = \frac{1}{|N_l|}\sum_l^{N_l} \left(\mathcal{L}_{SPM}^l + \mathcal{L}_{SFP}^l \right)
	\label{eqn-DD}
\end{equation}
where $N_l$ indicates the layer numbers of feature extractor network. Through the joint optimization of SPM and SFP, the model can resolve catastrophic forgetting and semantic drift with a mutually reinforced relationship.

\subsection{Uncertainty-aware Pseudo-labelling}
\label{Sec-UPL}
Self-supervised pseudo-labelling is a typical way to alleviate semantic drift since lack of old data. The pioneers~\cite{PLOP, IDEC} propose class-specific confidence-based threshold for pseudo labelling but ignore the specific challenging of dense prediction, i.e., the pixel-wise prediction needs pixel-wise optimization rather than class-wise threshold since the large intra-class variance. To enhance the anti-forgetting ability and reduce semantic drift during IL steps, this paper proposes an uncertainty-aware pseudo-labelling method. 

Particularly, the certainty score of pixel $i$ is defined by the absolute differentials between the highest prediction score $p_{i}^{h}$ and the second-highest score $p_{i}^{h'}$ predicted by $M^{t-1}$. 
\begin{equation}
	u_{i} = \vert p_{i}^{h}-p_{i}^{h'} \vert, i\in x, s.t., \mathcal{M}^{ct}=[u_{i}]_{H\times W}
\end{equation}	
where $\mathcal{M}^{ct}$ denotes the certainty-map. Thus the uncertainty of pixel $i$ is defined as $\vert 1-u_i \vert$, leading to the uncertainty-map $\mathcal{M}^{u}=\mathcal{M}_s^{t-1}\odot (\textbf{1}-\mathcal{M}^{ct})$, where $\textbf{1}$ is a matrix has the same size with $\mathcal{M}^{u}$ and all its elements are 1. $\odot$ denotes point-wise multiplication and $\mathcal{M}_s^{t-1}$ indicates the predicted soft segmentation map from $M^{t-1}$.
 A lower uncertainty indicates the prediction is more reliable. The protocol is to preserve the reliable predictions as pseudo labels since error supervision aggravates the model degradation. However, it is easily observed that the low-performance classes contribute the most degradation to the catastrophic forgetting. Here a self-adapted uncertainty-aware strategy is further proposed to enhance the incremental training efficiency based on the ranking of certainty. For pixel $i$, the uncertainty range $\triangle_i$ is:
 \begin{equation}
 	\triangle_i = |p_i^h-p_i^l|,  i\in x, p\in C^{0:t-1}
 \end{equation}  
where $p_i^l$ is the lowest prediction score on pixel $i$. Thus the pseudo label $\bar{y}_i^t$ for pixel $i$ is defined as:
\begin{equation}
	\begin{aligned}
		\bar{y}_i^t = \left\{ 
		\begin{aligned} 
			&\hat{y}_i^{t-1},  \rm{if} \ (p_i^h \in C^{1:t-1}) \land (p_i^h \geq \Gamma) \land  (u_i/\triangle_i \geq \zeta) \\
			&c^u, \quad \rm{if} \ (p_i^h\in c^b) \land (p_i^h <\Gamma) \land (u_i / \triangle_i < \zeta)  \\ 
			&c^b, \quad otherwise \\
		\end{aligned}
		\right. \\
	\end{aligned}
\end{equation}
where $\Gamma$ is the threshold for filtering the reliable predictions. $\zeta$ is used to judge the stability of the prediction. $\hat{y}_i^{t-1}$ is the prediction from $M^{t-1}$. Whereafter the supervision for $t$ step is composed by the element-wise addition by the pseudo labels and labels for current classes.

The integrated objective is defined as:
\begin{equation}
	\mathcal{L}_{total} = \mathcal{L}_{CE} + \mathcal{L}_{NSC} + \mathcal{L}_{DD}
\end{equation}
where $\mathcal{L}_{CE}$ indicates the conventional cross-entropy loss for pixel-level classification supervised by pseudo labels with retrievable labels of the incremental classes. Particularly, at the initial step, only $\mathcal{L}_{CE}$ is calculated, which is the same as the static training manner. While at each incremental step, $\mathcal{L}_{total}$ is calculated for the whole network training.

\section{Experimental Analysis} 	
\label{Sec-Experiment}
\subsection{Datasets and Protocols}
\label{Sec-Dataset}
\noindent\textbf{Datasets}. The experiments were conducted on class incremental within the same domain and on class\&domain incremental scene. The following datasets are introduced.
1) PASCAL VOC 2012~\cite{VOC2012} is a widely used benchmark for semantic segmentation. It consists of 10582 training images and 1449 images for validation with 20 semantic classes and an extra background class.  We evaluate our model on 15-5 (2 steps), 15-1 (6 steps), 5-3 (6 steps) and 10-1 (11 steps) settings. For example, 15-1 means initially learning 15 classes and learning additional 5 classes at another step. 15-1 indicates initially learning 15 classes and then learning the additional one class at each step for a total of another 5 steps. 
2) ADE20K~\cite{ADE} is a large-scale semantic segmentation dataset containing 150 classes that cover indoor and outdoor scenes. The dataset is split into 20210 training images and 2000 validation images. Compared to Pascal VOC 2012, ADE20K covers a wider variety of classes in natural scenes with more instances and categories. We evaluate our model on 100-50 (2 steps), 100-10 (6 steps), 50-50 (3 steps) and 100-5 (11 steps) settings. 
3) ISPRS~\cite{ISPRS} consists of two airborne image datasets including Postdam and Vaihingen. Postdam contains 38 images with 6000$\times$6000 size while Vaihingen contains 16 images. Semantic content in Postdam and Vaihingen has been classified manually into six land cover classes, namely: \textit{impervious surfaces}, \textit{building}, \textit{low vegetation}, \textit{tree}, \textit{car} and \textit{clutter}. We conduct the challenging class\&domain incremental experiments on Postdam$\rightarrow$ Vaihingen, i.e., the initial step training is on Postdam while the incremental steps are on Vaihingen. For training convenience, we partition each image into 600$\times$600 patches in sequential order.  Following previous works~\cite{Li2022ClassIncrementalLN, IDEC}, we choose to ignore the \textit{clutter} class in the training and validation process since it only accounts for very few pixel quantities and its unclear semantic scope.  We evaluate our model on 4-1 (2 steps), 2-3 (2 steps), 2-2-1 (3 steps) and 2-1 (4 steps) settings. Dataset details are shown in Table~\ref{table-dataset}. 
\begin{table}[h]
	\centering
	\caption{Details of CSS datasets and settings.} 
	\setlength{\tabcolsep}{0.2mm}{
		{\begin{tabular*}{0.48\textwidth}{@{\extracolsep{\fill}}cccccc@{}}
				\toprule[0.5mm]
				Dataset& Content &CSS setting &Class & Train-num. &Val-num.\\
				\midrule
				VOC 2012 &wild & class-incre.&20 &10582 &1449\\
				ADE20K&\makecell{indoor\& \\outdoor}&class-incre.&150 &20210 &2000 \\
				ISPRS &\makecell{remote\\sensing}&\makecell{class\& \\domain incre.}&5&\makecell{2400 (Pos.)+\\272 (Vai.)} & \makecell{1400 (Pos.)+\\256 (Vai.)} \\
				\bottomrule[0.5mm]
		\end{tabular*}}{}}	
	\label{table-dataset}
\end{table}\\
\noindent\textbf{Metrics}. We compute Pascal VOC mean intersection-over-union (mIoU)~\cite{VOC2012} for evaluation: 
\begin{equation}
	mIoU = \frac{TP}{TP+FP+FN}
\end{equation}
where TP, FP and FN are the numbers of true positive, false positive and false negative pixels, respectively. 
Specifically, the metric was computed after the last step T for the initial classes $C^{0: init}$, for the incremented classes $C^{init+1:T}$, and for all classes $C^{0:T}$ (all). \\
\noindent \textbf{Protocols}. Following~\cite{PLOP, IDEC}, there are two different incremental settings: \emph{disjoint} and \emph{overlapped}. In both settings, only the current classes $C^t$ are labelled and others are set as background. In the former, images at $t$ step only contain $C^{0:t-1} \cup C^t \cup C^{bg}$. While the latter contains $C^{0:t-1} \cup C^{t} \cup C^{t+1: T} \cup C^{bg}$, which is more realistic and challenging. In this study, we focus on \emph{overlapped} setting in our experiments. We report two baselines for reference, i.e., \emph{fine-tuning} on $C^{t}$,  and training on all classes \emph{offline}. The former is the lower bound and the latter can be regarded as the upper bound in CSS tasks. We argue that limited incremental data access meets more realistic CSS scenes. Thus we introduce a new metric, i.e., the data-ratio $\delta$, which is defined by the ratio of the amount of data used for incremental training steps to all available data, to reveal the robustness and learning efficiency of CSS models.

\begin{figure}[tbp]
	\centering
	\includegraphics[scale=0.48]{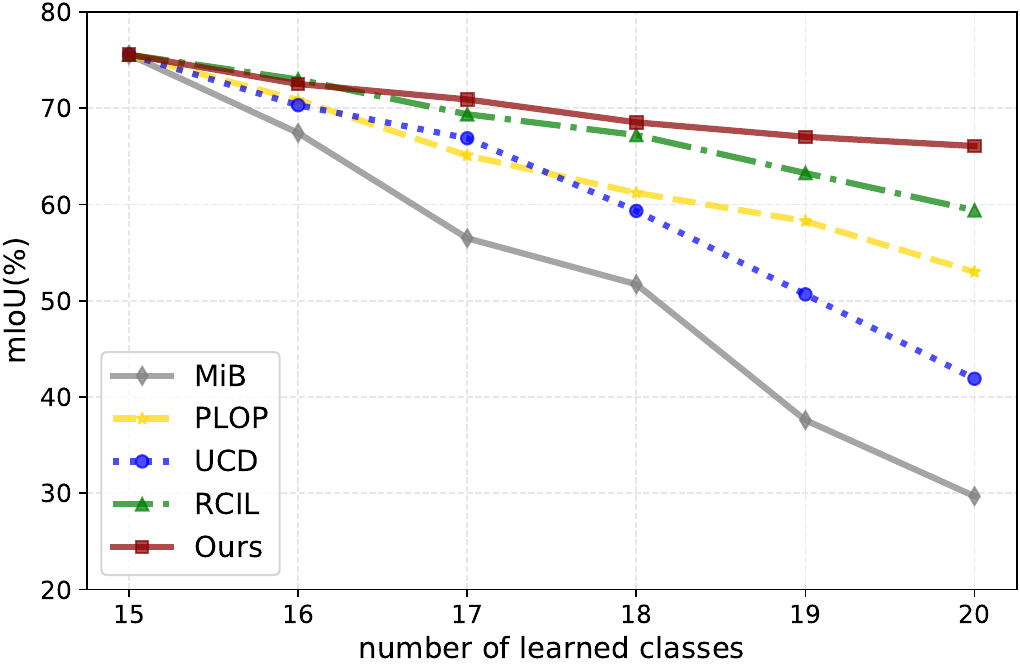}
	\caption{The mIoU (\%) evolution against number of learned classes on VOC 15-1 task.}
	\label{fig-mIoU_curve}
\end{figure}
\begin{figure}[tbp]
	\centering
	\includegraphics[scale=0.28]{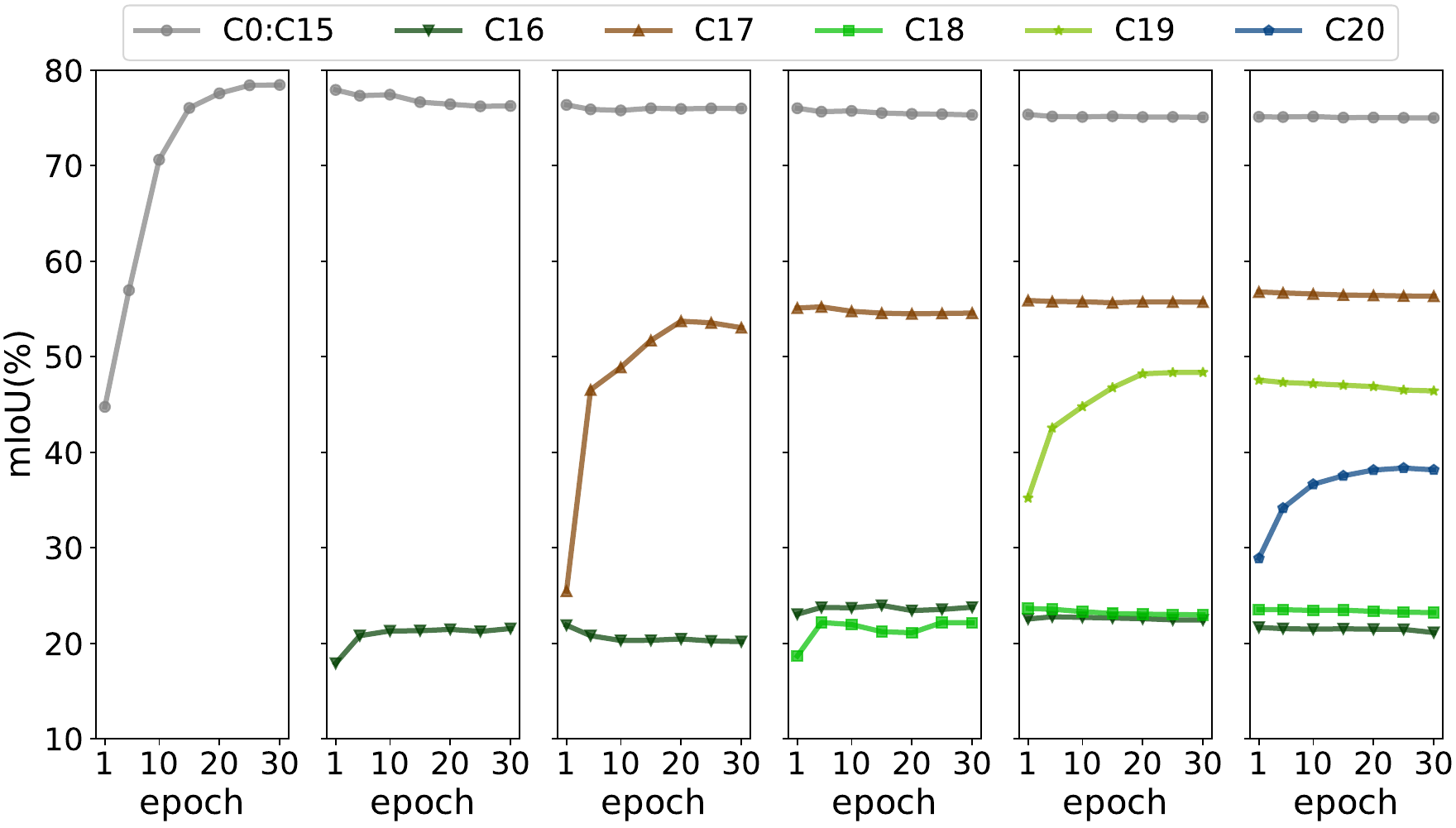}
	\caption{The mIoU (\%) evolution of the proposed model from Step0 to Step5 of learned classes and new classes on VOC 15-1 task.}
	\label{fig-mIoU_step}
\end{figure}

\subsection{Implementation Details}
For fair comparison, we implement the proposed model based on DeepLabv3~\cite{DeepLabv3} with ResNet-101~\cite{ResNet} as backbone. For all experiments, the initial learning rate is 0.01 and decayed by a \textit{poly} policy. The model was trained with 30 (VOC 2012,  ISPRS) and 50 (ADE20K) epochs for each IL step, respectively. The hyper-parameters are set as $\Gamma=0.7$ and $\rho=0.6$ according to the ablation study in Sec.~\ref{Sec-Abla-parameters}. $\zeta$ is set to 5. The input image is resized to 513$\times$513 and the training batch size is set to 24 in all CSS settings. Our implementation is based on PyTorch 1.8 with CUDA 11.6 and all experiments are conducted on a workstation with two NVIDIA A800 GPUs.  The code and models are available at \url{https://github.com/YBIO/LAG}.
It is worth mentioning the proposed architecture is designed to be model-agnostic. The impact of semantic segmentation models and backbones is explored in Sec.~\ref{Sec-Abla-SegModel}.

\subsection{Quantitative Evaluation}
\label{Sec-QE}
\begin{figure}[tbp]
	\centering
	\includegraphics[scale=0.7]{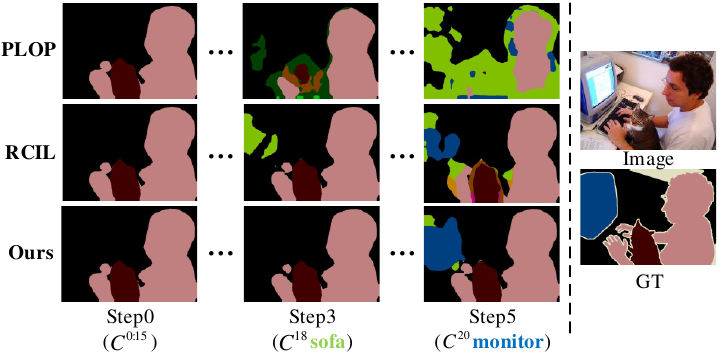}
	\caption{Visualization of our model predictions in comparison with PLOP~\cite{PLOP} and RCIL~\cite{RCIL} at incremental steps. At \textit{Step3} and \textit{Step5}, the class \textit{sofa} and \textit{monitor} are introduced, respectively. PLOP and  RCIL are plagued by background semantic drift. Our method shows superior robustness and valid compatibility on new classes.}
	\label{fig-vis-step}
\end{figure}

\noindent{\textbf{Class-incremental in VOC}}. We firstly evaluate the proposed model on class-incremental semantic segmentation task in VOC 2012. Following protocols in Sec.~\ref{Sec-Dataset}, the quantitative comparisons of our model with the representative pioneers including~\cite{EWC, LWF, MiB, PLOP, SDR, UCD, REMINDER, RCIL} are shown in Table~\ref{table-VOC2012}. On the one hand, the proposed model achieves superior performance at multi-step CSS tasks. In the challenging 10-1 task, where multiple steps of learning are involved and significant semantic drift occurs during IL steps. As a result, the pioneers normally suffer from model degradation in both old and new classes. As a contrast, the proposed model achieves higher mIoU performance in both initial learned classes and new learned classes, validating the effectiveness of the proposed method. On the other hand, the proposed model shows robust anti-forgetting performance, as shown in Fig.~\ref{fig-mIoU_curve}. Compared with current SOTA~\cite{PLOP, RCIL}, LAG maintains a high performance in mIoU, indicating the compatibility in old knowledge inheritance and new class learning. However, LAG performs inferior to RCIL in 15-5 task, which to our knowledge, is the semantic confusion in class-wise prototype matching when leaning multiple novel classes.
Fig.~\ref{fig-mIoU_step} shows the mIoU evolution of the initial $C^{0:15}$ and each new learned class in terms of VOC 15-1. It shows the learned classes faces catastrophic forgetting at IL steps such as $C^{18}$ (\textit{sofa}) and $C^{19}$ (\textit{train}). Interestingly, it also indicates the old classes may achieve better performance with the IL steps increasing.  For example, the IoU of $C^{17}$ (\textit{sheep}) at Step3-Step5 is higher than that at Step2. To our knowledge, it can be attributed to the gradual dismissal of semantic confusion with the background as more classes are learned.
Fig.~\ref{fig-vis-step} illustrates the comparative visualized performance, it also demonstrates LAG's superior ability in alleviating catastrophic forgetting and semantic drift. 
\begin{table*}[htbp]
	\caption{Quantitative comparison on VOC 2012 in terms of mIoU (\%).  The segmentation model is DeepLabv3~\cite{DeepLabv3}. Class 0 indicates the unlabelled class. $\ddagger$ means we re-implement the method, and the others are from the original paper. The first and second best results are highlighted in bold and underlined, respectively.}
	\centering
	\setlength{\tabcolsep}{1mm}{
		{\begin{tabular*}{0.95\textwidth}{@{\extracolsep{\fill}}l|ccc|ccc|ccc|ccc@{}}
				\toprule[0.5mm]
				\multirow{2}*{Method} &
				\multicolumn{3}{c}{15-5 (2 steps)} & \multicolumn{3}{c}{15-1 (6 steps)} & \multicolumn{3}{c}{5-3 (6 steps)} & \multicolumn{3}{c}{10-1 (11 steps)} \\
				& 0-15 & 16-20 & all & 0-15 & 16-20 & all & 0-5 & 6-20 & all & 0-10 & 11-20 & all\\
				\midrule
				\emph{fine tuning}&2.10&33.10&9.80&0.20&1.80&0.60&0.50&10.40&7.60&6.30&2.80&4.70\\
				EWC~\cite{EWC} &24.30 &35.50 &27.10 &0.30 &4.30 &1.30 &-&-&-&- &- &-\\
				LwF-MC~\cite{LWF} &58.10&35.00&52.30&6.40&8.40&6.90&20.91&36.67&24.66&4.65&5.90&4.95\\
				ILT~\cite{ILT} &66.30 &40.60&59.90&4.90&7.80&5.70&22.51&31.66&29.04&7.15&3.67&5.50\\
				MiB~\cite{MiB}&76.37&49.97&70.08&34.22&13.50&29.29&57.10&42.56&46.71&12.25&13.09&12.65\\
				PLOP~\cite{PLOP}&75.73&51.71&70.09&65.12&21.11&54.64&17.48&19.16&18.68&44.03&15.51&30.45 \\
				SDR~\cite{SDR} &75.40&\textbf{52.60}&69.90&44.70&21.80&39.20&-&-&-&32.40&17.10&25.10 \\
				UCD+PLOP~\cite{UCD}&75.00&51.80&69.20&66.30&21.60&55.10&-&-&-&42.30&\underline{28.30}&\underline{35.30}\\
				REMINDER~\cite{REMINDER} &76.11&50.74&70.07&68.30&\underline{27.23}&58.52&-&-&-&-&-&- \\
				RCIL~\cite{RCIL}&\textbf{78.80}&\underline{52.00}&\textbf{72.40}&\underline{70.60}&23.70&\underline{59.40}&\underline{65.30}&\underline{41.49}&\underline{50.27}&\underline{55.40}&15.10&34.30\\
				\cmidrule{1-13} 	
				\textbf{Ours} &\underline{77.33}&51.76&\underline{71.24}&\textbf{75.00}&\textbf{37.52}&\textbf{66.08}&\textbf{67.53}&\textbf{47.11}&\textbf{52.94}&\textbf{69.56}&\textbf{42.62}&\textbf{56.73}\\
				\midrule
				\emph{offline} &79.77&72.35&77.43&79.77&72.35&77.43&76.91&77.63&77.43&78.41&76.35&77.43\\
				\bottomrule[0.5mm]
		\end{tabular*}}{}}	
	\label{table-VOC2012}
\end{table*}

\noindent{\textbf{Class-incremental in ADE}}. The class-incremental semantic segmentation performance on ADE20K is shown in Table~\ref{table-ADE20K}. We also reported the model performance of the representative pioneers and our model under four IL settings. Since large-amount class with the abundant semantics, the performance on this dataset is low and indicates the severe semantic confusion. For example, in the challenging 100-5 task, the pioneers~\cite{MiB, PLOP} degrades drastically in old classes and is not quite adaptive to new classes after IL steps. Current state-of-the-art~\cite{RCIL} also faces low performance in the new classes. The proposed LAG achieves a better compatibility in both anti-forgetting on old classes and the adaptation on new classes. Concretely, LAG  achieves 39.96\% in $C^{1:100}$ and 17.22\% mIoU in $C^{101:150}$, respectively. To our knowledge, the CSS performance is also affected by the effectiveness of semantic segmentation models, which is explored in Sec.~\ref{Sec-Abla-SegModel}. 
\begin{table*}[htbp]
	\caption{Quantitative comparison on ADE20K in terms of mIoU (\%). The segmentation model is DeepLabv3~\cite{DeepLabv3}.  The first and second best results are highlighted in bold and underlined, respectively.}
	\centering
	\setlength{\tabcolsep}{1mm}{
		{\begin{tabular*}{0.95\textwidth}{@{\extracolsep{\fill}}l|ccc|ccc|ccc|ccc@{}}
				\toprule[0.5mm]
				\multirow{2}*{Method} &
				\multicolumn{3}{c}{100-50 (2 steps)}   & \multicolumn{3}{c}{100-10 (6 steps)} & \multicolumn{3}{c}{50-50 (3 steps)} & \multicolumn{3}{c}{100-5 (11 steps)}\\
				& 1-100 & 101-150 & all & 1-100 & 101-150 & all & 1-50 & 51-150 & all  & 1-100 & 101-150 & all\\
				\midrule
				\emph{fine tuning}&0.00&11.22&3.74&0.00&2.08&0.69&0.00&3.60&2.40&0.00&0.07&0.02\\
				ILT~\cite{ILT} &18.29&14.40&17.00&0.11&3.06&1.09 &3.53&12.85&9.70&0.08&1.31&0.49\\
				MiB~\cite{MiB}&40.52&17.17&32.79&38.21&11.12&29.24&45.57&21.01&29.31&36.01&5.66&25.96\\
				PLOP~\cite{PLOP}&41.87&14.89&32.94&40.48&13.61&31.59&\textbf{48.83}&20.99&30.40&\underline{39.11}&7.81&28.75\\
				UCD+PLOP~\cite{UCD}&\underline{42.12}&15.84&33.31&\underline{40.80}&15.23&32.29&47.12&24.12&31.79&-&-&-\\
				REMINDER~\cite{REMINDER} &41.55&\underline{19.16}&34.14&38.96&\textbf{21.28}&\underline{33.11}&47.11&20.35&29.39&36.06&\underline{16.38}&29.54 \\
				RCIL~\cite{RCIL}&\textbf{42.30}&18.80&\textbf{34.50}&39.30&17.60&32.10&\underline{48.30}&\underline{25.00}&\underline{32.50}&38.50&11.50&\underline{29.60}\\
				\cmidrule{1-13} 
				\textbf{Ours} &41.64&\textbf{19.73}&\underline{34.34}&\textbf{41.00}&\underline{18.69}&\textbf{33.56}&47.69&\textbf{26.12}&\textbf{33.31}&\textbf{39.96}&\textbf{17.22}&\textbf{32.38}\\
				\midrule
				\emph{offline} &44.30&28.20&38.90&44.30&28.20&38.90&50.90&32.90&38.90&44.30&28.20&38.90\\
				\bottomrule[0.5mm]
		\end{tabular*}}{}}
	\label{table-ADE20K}
\end{table*}
\begin{figure}[tbp]
	\centering
	\includegraphics[scale=0.73]{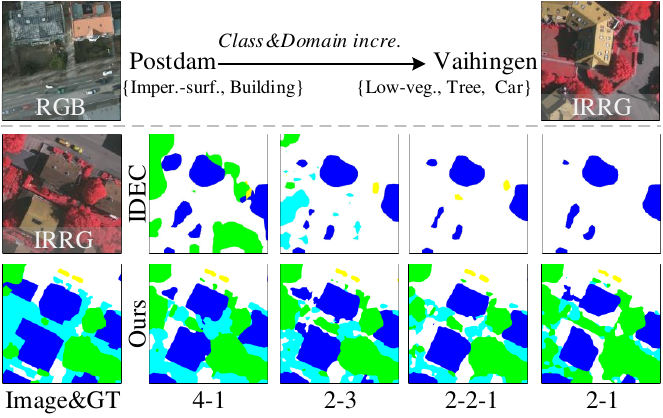}
	\caption{Illustration of class\&domain incremental semantic segmentation results from Postdam (RGB)$\rightarrow$Vaihingen (IRRG). Compared with the pioneer~\cite{IDEC}, the proposed model achieves superior IL performance on both the old and new classes.}
	\label{fig-vis-ISPRS}
\end{figure}

\begin{figure*}[tbp]
	\centering
	\includegraphics[scale=0.73]{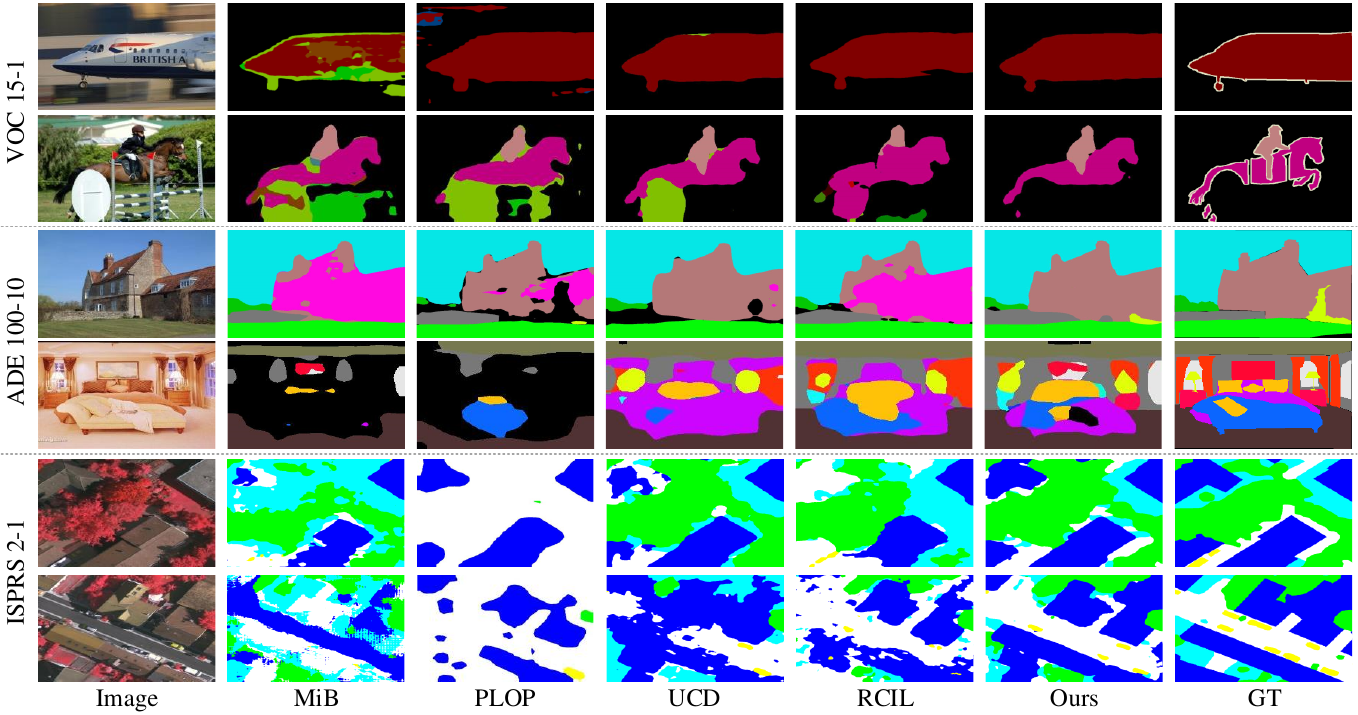}
	\caption{Qualitative visualizations from various approaches on VOC 15-1 (class incre.), ADE 100-10 (class incre.) and ISPRS 2-1 (class\&domain incre.) tasks. The illustration demonstrates the superiority of our approach on both new and old classes.}
	\label{fig-vis-comparison}
\end{figure*}

\noindent{\textbf{Class\&domain-incremental in ISPRS}}. 
To meet realistic applications, we conduct a class\&domain incremental setting on ISPRS as introduced in Sec.~\ref{Sec-Dataset}. Particularly, images in Postdam are RGB format while IRRG format in Vaihingen, leading to a significant domain variance. Note that for training, the initial training step is on Postdam for base classes and the IL steps are on Vaihingen for incremental classes. While for testing, the validation set consists of samples from both Postdam and Vaihingen to evaluate the anti-forgetting and compatibility on new classes simultaneously. Table~\ref{table-ISPRS} shows the performance comparison between the proposed LAG with the state-of-the-arts. The large diversity between the incremental and base classes and the limitation of incremental data in Vaihingen indicate that such a task is a huge challenge, which is revealed by the model degradation by~\cite{RCIL, IDEC}. Compared to feature-replay-based method IDEC+PM, the proposed LAG also achieves superior performance in both anti-forgetting on old classes and adaptability on incremental classes. Concretely, LAG achieves the highest mIoU on all classes after the final step on all four CSS settings, especially in the most challenging ISPRS 2-1 task with 48.67\% mIoU. It also proves that feature-replay integrates alleviating catastrophic forgetting and avoiding storage burdens and privacy concerns. Note that the performance of \textit{offline} setting is based on the joint training on both Postdam and Vaihingen datasets. Fig.~\ref{fig-vis-ISPRS} illustrates the visualized performance in such a class\&domain incremental task, demonstrating the sever semantic drift in~\cite{IDEC} and the better compatibility of the proposed model.  The qualitative comparisons across three datasets are shown in Fig.~\ref{fig-vis-comparison}, which demonstrate the proposed model achieves solid semantic perception in various CSS scenarios.
\begin{table*}[!htbp]
	\centering
	\footnotesize
	\caption{Quantitative comparison on ISPRS in mIoU (\%).  The first and second best results are highlighted in bold and underlined, respectively. $\ddagger$ means we re-implement the method under current CSS settings. Note that the validation set contains both Postdam and Vaihingen images to evaluate the initial classes\&domains and the incremental classes\&domains simultaneously.}
	\setlength{\tabcolsep}{1mm}{
		{\begin{tabular*}{0.95\textwidth}{@{\extracolsep{\fill}}l|ccc|ccc|ccc|ccc@{}}
				\toprule[0.5mm]
				\multirow{2}*{Method} &
				\multicolumn{3}{c}{4-1 (2 steps)}   & \multicolumn{3}{c}{2-3 (2 steps)} & \multicolumn{3}{c}{2-2-1 (3 steps)} & \multicolumn{3}{c}{2-1 (4 steps)}\\
				& 1-4 & 5 & all & 1-2 & 3-5 & all & 1-2 & 3-5 & all  & 1-2 & 3-5 & all\\
				\midrule
				\emph{fine tuning}&2.44 &9.29 &3.81 &0.72 &5.51 &3.59 &0.09 &3.87 &2.36 &0.03 &4.27 &2.57\\
				MiB$^\ddagger$~\cite{MiB}&22.70&11.25&20.41&28.94&7.11&15.84&22.58&5.99&12.63&27.21&4.03&13.30\\
				PLOP$^\ddagger$~\cite{PLOP}&35.88&7.30&30.16&37.00&5.91&18.35&29.07&8.22&16.56&33.82&3.98&15.92 \\
				RCIL$^\ddagger$~\cite{RCIL} &40.26&11.33 &34.47 &47.22 &9.43 &24.55 &39.88 &6.83 &20.05 &35.29 &4.99 &17.11 \\
				IDEC$^\ddagger$~\cite{IDEC}&38.74&15.07&34.00&59.26&10.78&30.17&46.45&5.95&22.15&46.87&5.42&22.00\\
				IDEC+PM$^\ddagger$~\cite{IDEC}&\underline{61.45}&\underline{32.28}&\underline{55.62}&\underline{62.70}&\underline{41.01}&\underline{49.69}&\underline{58.25}&\underline{32.00}&\underline{42.50}&\underline{59.75}&\textbf{35.63}&\underline{45.28}\\
				\midrule
				\textbf{Ours}&\textbf{66.60}&\textbf{38.00}&\textbf{60.88}&\textbf{72.55}&\textbf{43.77}&\textbf{55.28}&\textbf{68.74}&\textbf{35.30}&\textbf{48.68}&\textbf{68.78}&\underline{35.27}&\textbf{48.67}\\
				\midrule
				\emph{offline} &74.10&63.21&71.92&82.15&65.11&71.92&82.15&65.11&71.92&82.15&65.11&71.92\\
				\bottomrule[0.5mm]
		\end{tabular*}}{}}	
	\label{table-ISPRS}
\end{table*}

\subsection{Ablation Study}
\subsubsection{Module Contribution}
\label{Sec-Abla-MC}
To reveal the contribution of each module in the proposed method, we respectively disclose the corresponding modules. In Table~\ref{table-Aba-Module}, on the one hand, the proposed disentangled distillation validates its effectiveness through the coherence of SPM module and SFP mechanism. Concretely, it is observed that on VOC 15-1 task, the collaboration of SPM and SFP can effectively boost the anti-forgetting on old classes $C^{0:15}$ and learning efficiency on new classes $C^{16:20}$ synchronously. In addition, it also proves the prototypes matching is an effective means in CSS tasks. The improvement confirms our intuition of disentangled distillation on solid knowledge inheritance.  On the other hand, the NSC constraint brings diverse boosting on learned classes and new classes, and the overall performance can be improved due to the consistency in latent space.
\begin{table}[t]
	\centering
	\caption{Individual contribution of proposed modules. The results come from VOC 15-1, ADE 100-10 and ISPRS 2-1 tasks.}
	\setlength{\tabcolsep}{0.45mm}{
		\begin{tabular}{l|ccc|ccc|ccc}
			\toprule[0.4mm]
			\multirow{2}*{Method} &\multicolumn{3}{c}{VOC 15-1} & \multicolumn{3}{c}{ADE 100-10} &\multicolumn{3}{c}{ISPRS 2-1}  \\
			& 0-15 & 16-20 & all & 1-100 & 101-150 & all &1-2&3-5&all \\
			\midrule
			\emph{fine tuning} &0.20&1.80&0.60&0.00&2.08&0.69 &0.89&1.53&1.27\\
			\midrule
			+SPM (Eqn.~\ref{eqn-SPM})&71.35&32.97&62.21&37.42&14.56&29.80&66.22&32.58&46.04\\
			+SFP (Eqn.~\ref{eqn-SFP})&69.11&33.12&60.54&36.53&14.00&29.02&42.77&27.44&33.57\\
			+SPM\&SFP &74.26&\textbf{37.69}&65.55&\textbf{41.21}&18.17&33.53&\textbf{68.86}&33.25&47.49\\	
			\midrule
			+NSC (Eqn.~\ref{eqn-NSC})&\textbf{75.00}&37.52&\textbf{66.08}&41.00&\textbf{18.69}&\textbf{33.56}&68.78&\textbf{35.27}&\textbf{48.67}\\
			\bottomrule[0.5mm]
	\end{tabular}}
	\label{table-Aba-Module}
\end{table}
\begin{table}[t]
	\centering
	\caption{Performance comparison using various pseudo-labelling methods including \textit{None} threshold, \textit{fixed} threshold (set as 0.7), class-specific threshold from~\cite{PLOP} and~\cite{IDEC}, as well as the proposed UPL.} 
	\setlength{\tabcolsep}{1.0mm}{
	\begin{tabular}{l|ccccc|c}
		\toprule[0.5mm]
		Task&\textit{None}&\textit{fixed}&PLOP~\cite{PLOP}&DCPL~\cite{IDEC}&UPL&mIoU(\%)\\
		\midrule
		\multirow{5}*{\makecell{VOC\\15-1}}&\checkmark&&&&&59.71\\
		&&\checkmark&&&&63.79\\
		&&&\checkmark&&&61.52\\
		&&&&\checkmark&&65.67\\
		&&&&&\checkmark&\textbf{66.08}\\
		\midrule
		\multirow{5}*{\makecell{ADE\\100-10}}&\checkmark&&&&&29.50\\
		&&\checkmark&&&&30.51\\
		&&&\checkmark&&&31.99\\
		&&&&\checkmark&&\textbf{33.68}\\
		&&&&&\checkmark&33.56\\
		\midrule
		\multirow{5}*{\makecell{ISPRS\\2-1}}&\checkmark&&&&&35.18\\
		&&\checkmark&&&&46.84\\
		&&&\checkmark&&&43.61\\
		&&&&\checkmark&&48.15\\
		&&&&&\checkmark&\textbf{48.67}\\
		\bottomrule[0.5mm]
	\end{tabular}}
	\label{table-Aba-Pseudo}
\end{table}

\subsubsection{Impact of Pseudo-labelling}
Pseudo-labelling is an effective way to alleviate catastrophic forgetting due to the lack of old data and annotation. We compare the proposed UPL with the state-of-the-art class-specific pseudo-labelling methods in~\cite{PLOP} and~\cite{IDEC}. The quantitative comparison is performed based on ADE 100-10 and ISPRS 2-1 tasks. The former indicates multi-class indoor and outdoor scenes. The latter indicates the class\&domain incremental remote-sensing scene. As seen in Table~\ref{table-Aba-Pseudo}, the proposed UPL achieves the highest performance on VOC 15-1 and ISPRS 2-1, which validates the effectiveness of UPL. However, on ADE 100-10, the class-specific pseudo-labelling method~\cite{IDEC} is more superior, which indicates the large-amount class number and complex inter-class with intra-class variance is the key of achieving efficient pseudo-labelling.

\begin{table}[htbp]
	\centering
	\caption{Impact of $\rho$ on VOC 15-1 task.} 
	\setlength{\tabcolsep}{1.0mm}{
		\begin{tabular}{c|ccccccc}
			\toprule[0.5mm]
			$\rho$&0&0.3&0.5&0.6&0.7&0.9&1.0\\
			\midrule
			mIoU (\%)&57.48&62.39&65.58&\textbf{66.08}&64.92&62.58&60.44\\			
			\bottomrule[0.4mm]
	\end{tabular}}
	\label{table-Aba-rho}
\end{table}	  
\begin{table}[htbp]
	\centering
	\caption{Impact of $\Gamma$ on VOC 15-1 task.}
	\setlength{\tabcolsep}{1.0mm}{
		\begin{tabular}{c|cccccc}
			\toprule[0.4mm]
			$\Gamma$&0&0.5&0.6&0.7&0.8&0.9\\
			\midrule
			mIoU (\%)&49.37&52.22&65.01&\textbf{66.08}&62.52&59.88\\			
			\bottomrule[0.5mm]
	\end{tabular}}
	\label{table-Aba-Gamma}
\end{table}
\subsubsection{Impact of Hyper-parameters}
\label{Sec-Abla-parameters}
 There are several hyper-parameters in the proposed network including $\rho$ and $\Gamma$ introduced in Sec.~\ref{Sec-NSC} and Sec.~\ref{Sec-UPL}, respectively.  The impact of parameter settings on the final performance is explored in this section. As seen in Table~\ref{table-Aba-rho}, the disentangling of semantic-invariant term and sample-specific term can maximally boost the performance by 8.60\% mIoU improvement than the naive distillation mechanism introduced in Eqn.~\ref{eqn-SFP}, i.e., $\rho=0$. It also proves our intuition about semantic-invariance modelling, and validates the collaboration of the proposed SPM and SFP. Table~\ref{table-Aba-Gamma} reveals the impact of $\Gamma$ in controlling confidence of pseudo labels. Note these hyper-parameters are not designed for a specific dataset, despite we only present the quantitative results on VOC 2012.

\begin{figure*}[htbp]
	\centering
	\includegraphics[scale=0.48]{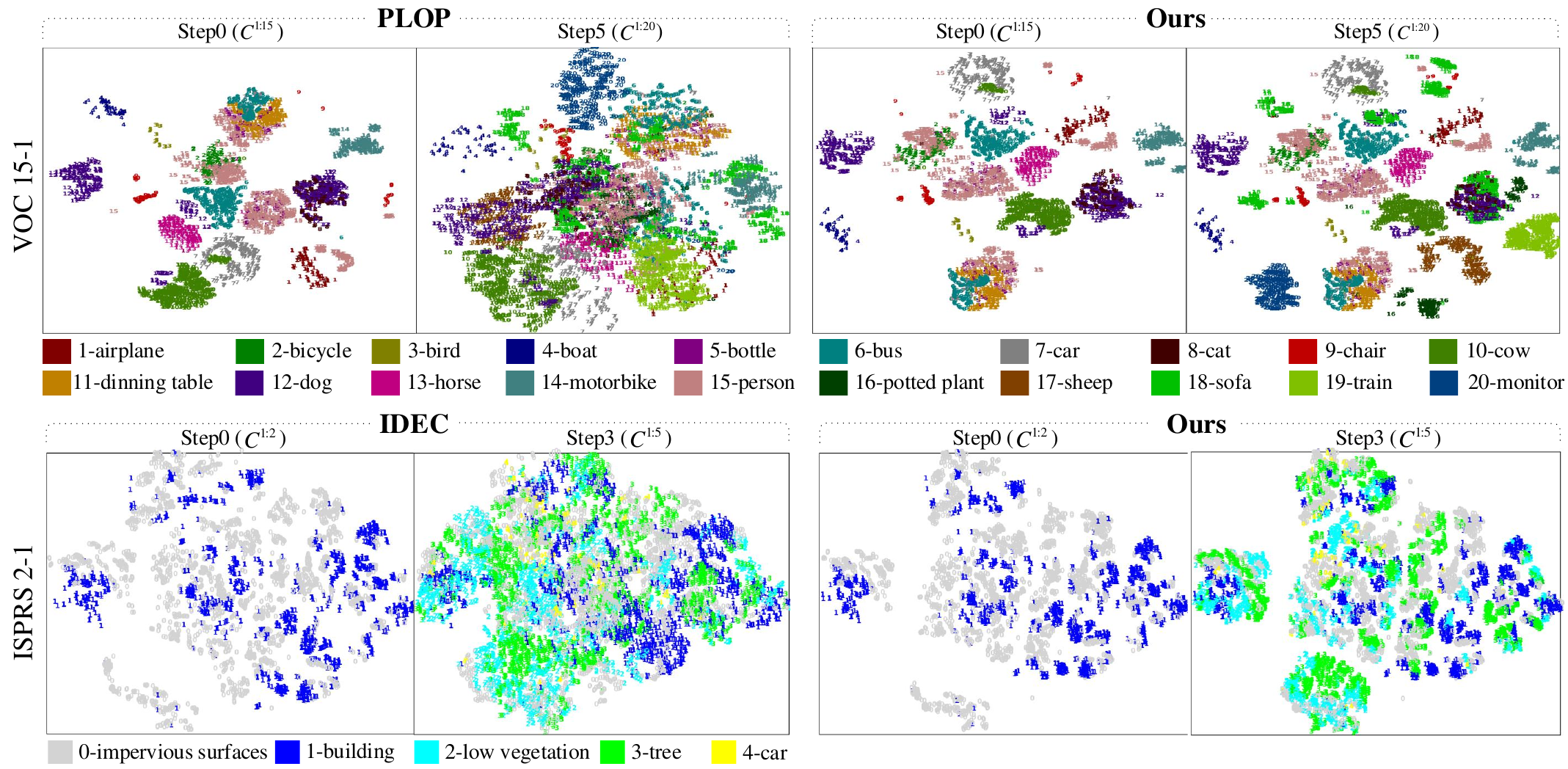}
	\caption{T-SNE visualizations with respect to VOC 15-1 and ISPRS 2-1 at the initial step and the final step, best viewed in colour. For visualization convenience, the \textit{bg} class in VOC 2012 is skipped. Numbers in the image represent the corresponding classes.}
	\label{fig-tsne}
\end{figure*}

\subsubsection{Impact of Segmentation Model}
\label{Sec-Abla-SegModel}
The proposed method is model-agnostic, i.e., it is adaptable to various semantic segmentation models and backbones. Table~\ref{table-Abla-segmodel} explores the impact the CSS performance with various semantic segmentation models~\cite{DeepLabv3, DeepLabv3+} and backbones~\cite{ResNet, SwinTH} on VOC 15-1, ADE 100-10 and ISPRS 2-1 tasks.  On the one hand, the impact of the encoder by comparing CNN and the newfangled Transformer architecture is explored. For example, DeepLabv3 with Swin-T achieves superior performance on the incremental classes and leads to better mIoU on all classes than that with ResNet-101. It indicates that a stronger backbone is able to maintain a higher performance after IL steps. Note that the NSC constraint defined in Eqn.~\ref{eqn-NSC} is not applied in Transformer architectures.  On the other hand, compared with DeepLabv3, DeepLabv3+ model achieves 0.87\% and 0.50\% mIoU improvements on the initial $C^{1:100}$ and $C^{all}$ respectively in terms of ADE 100-10. Peculiarly, it is observed that on the challenging class\&domain incremental task ISPRS 2-1, DeepLabv3 with ResNet-101 is superior to that with Swin-T, which mainly result from the large data variance and limited data quantity for training.

\begin{table}[tbp]
	\caption{CSS performance using various semantic segmentation models and backbones.}
	\centering
	\begin{tabular}{c|l|c|ccc}
		\toprule[0.5mm]
		Task&\makecell[c]{Model} &Backbone&$C^{init.}$&$C^{incre.}$&all\\
		\midrule
		\multirow{4}*{\makecell[c]{VOC\\15-1}}&DeepLabv3&ResNet-101&75.00&37.52&66.08\\
		&DeepLabv3&Swin-T&75.31&37.79&66.38\\
		&DeepLabv3+&ResNet-101&74.55&38.27&65.91\\
		&DeepLabv3+&Swin-T&74.99&39.12&65.26\\
		\midrule
		\multirow{4}*{\makecell[c]{ADE\\100-10}}&DeepLabv3&ResNet-101&41.00&18.69&33.56\\
		&DeepLabv3&Swin-T&41.14&17.88&33.39\\
		&DeepLabv3+&ResNet-101&41.87&18.43&34.06\\
		&DeepLabv3+&Swin-T&41.45&19.02&33.97\\
		\midrule
		\multirow{4}*{\makecell[c]{ISPRS\\2-1}}&DeepLabv3&ResNet-101&68.78&35.27&48.67\\
		&DeepLabv3&Swin-T&67.29&29.43&44.57\\
		&DeepLabv3+&ResNet-101&69.22&37.31&50.07\\
		&DeepLabv3+&Swin-T&68.41&35.23&48.50 \\
		\bottomrule[0.5mm]
	\end{tabular}
	\label{table-Abla-segmodel}
\end{table}

\subsubsection{Interpretability Analysis}

To investigate the learning efficiency upon inner feature distribution at incremental steps, we use t-SNE~\cite{TSNE} to map the high-dimensional features to 2D space. As illustrated in Fig.~\ref{fig-tsne}, we compare the feature distribution between PLOP~\cite{PLOP} and our method at the initial step including $C^{0:15}$ and the final step including $C^{0:20}$ in terms of VOC 15-1. On the one hand, it substantiates the classifier bias after incremental steps since PLOP is up against classifier failure in the latent space at Step-5.  On the other hand, the illustration demonstrates our method achieves balancing of both plasticity and stability, which is revealed by the solid anti-forgetting on old classes and adaptation to new classes. We also compare the ISPRS 2-1 performance between IDEC~\cite{IDEC} and our model. It demonstrates the severe catastrophic forgetting and semantic drift align with the quantitative metric in Table~\ref{table-ISPRS}, which also validates the superiority of the proposed model on class\&domain incremental task.

\begin{figure}[tbp]
	\centering
	\includegraphics[scale=0.43]{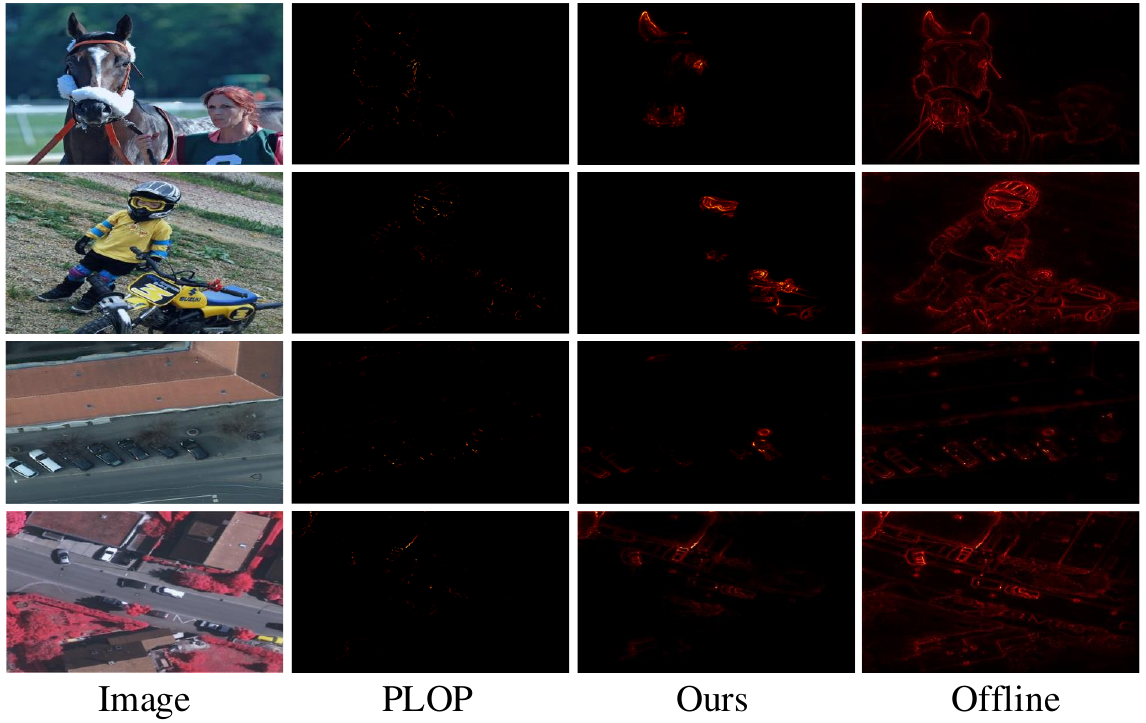}
	\caption{Interpretability map based on LRP~\cite{LRP}.}
	\label{fig-LRP}
\end{figure}

The semantic-invariance modelling method introduced in Sec.~\ref{Sec-NSC} is able to boost the feature alignment during IL steps. As shown in Fig.~\ref{fig-LRP}, the gradient response at the final step indicates the model tends to discard the sensitivity of the previous learned classes. While the proposed model alleviates this degeneration.

\subsubsection{Robustness to Limited Incremental Data}
\begin{table}[!tbp]
	\centering
	\caption{The sample number for training at incremental steps under data-limited conditions on VOC 15-1 task.}
	\setlength{\tabcolsep}{1.5mm}{
		\begin{tabular}{l|ccccc}
			\toprule[0.5mm]
			Data Limitation&$\delta$=100\%&$\delta$=75\%&$\delta$=50\%&$\delta$=25\%&$\delta$=10\%\\
			\midrule
			Step0 ($C^{0:15}$)&9568&9568&9568&9568&9568\\
			\midrule
			Step1 ($C^{16}$)&487&365&244&122&49\\
			Step2 ($C^{17}$)&299&224&150&75&30\\
			Step3 ($C^{18}$)&491&368&246&123&49\\
			Step4 ($C^{19}$)&500&375&250&125& 50\\
			Step5 ($C^{20}$)&548&411&274&137&55\\
			\bottomrule[0.5mm]
	\end{tabular}}
	\label{table-limitdata}
\end{table}

\begin{table}[!tbp]
	\centering
	\caption{Ablation study of limited incremental data condition on VOC 15-1.}
	\setlength{\tabcolsep}{0.5mm}{
		\begin{tabular}{l|ccc|ccc}
			\toprule[0.5mm]
			\multirow{2}*{$\delta$}&\multicolumn{3}{c|}{PLOP}&\multicolumn{3}{c}{Ours}  \\
			&0-15&16-20&all&0-15&16-20&all\\
			\midrule
			\rowcolor{lightgray}100\%&65.12&21.11&54.64&75.00&37.52&66.08\\
			75\%&64.45&15.32&52.75&74.01&36.72&65.14\\
			50\%&62.75&13.46&51.01&74.12&35.58&64.94\\
			25\%&60.11&9.79&48.13&74.40&31.28&64.13\\
			\rowcolor{lightgray}10\%&51.26&5.38&40.34&74.31&15.54&60.32\\
			\midrule
			\rowcolor{lightgray} $gain$&$\downarrow 21.28\%$&$\downarrow 74.51\%$&$\downarrow 26.17\%$&$\downarrow 0.92\%$&$\downarrow 58.58\%$&$\downarrow 8.72\%$\\
			\bottomrule[0.5mm]
	\end{tabular}}
	\label{table-Aba-Datalimitation}
\end{table}
Due to the lack of new samples and the difficulty of data annotation, the incremental training with limited data is more in line with practical application. Following the protocol in Sec.~\ref{Sec-Dataset}, at each incremental step, the sample quantity of new classes is constrained with various ratios from full data to partial data for training. In detail, we organize the data as seen in Table~\ref{table-limitdata}. The training data at the initial step, i.e., Step0 is not restrained. At Step1, the total image quantity for the incremental class $C^{16}$ (\textit{potted plant}) is 487. Under the data-limited condition, we sample images following the sequence order in the dataset. For example, with respect to $\delta$=10\%, the training image number of $C^{16}$ decreases to 49, which brings a great challenge for the model to learn the new class. 

The limited data condition is able to reveal the generalization and robustness of the model. Specifically, we re-evaluate VOC 15-1 as seen in Table~\ref{table-Aba-Datalimitation}. The representative pioneer~\cite{PLOP} degenerates drastically when incremental training data is limited. For example, PLOP nearly fails to adapt to new classes with 5.38\% mIoU on $C^{16:20}$ when $\delta$ is limited to 10\%. As a comparison, the proposed method achieves more superior performance under the data-limited conditions. Particularly, LAG maintains 95\% performance (35.58 vs. 37.52) on the incremental $C^{16:20}$ when $\delta=50\%$. In the most challenging $\delta=10\%$ setting, the proposed model  achieves 15.54\% mIoU on the incremental classes, while PLOP degrades to 5.38\%. On the other hand, LAG maintains 74.31\% mIoU on the initial learned $C^{0:15}$, showing stronger anti-forgetting on old classes. 

\subsubsection{Robustness to Class Incremental Orders}
To reveal the robustness to class orders of the proposed method, we perform experiments on VOC 15-1 with five different class orders including the ascending order and four random orders as follows.
\begin{equation}
	\scriptsize
	\begin{split}
		\nonumber
		a:\{[0,1,2,3,4,5,6,7,8,9,10,11,12,13,14,15],[16],[17],[18],[19],[20]\} \\
		b:\{[0,12,9,20,7,15,8,14,16,5,19,4,1,13,2,11],[17],[3],[6],[18],[10]\} \\
		c:\{[0,13,19,15,17,9,8,5,20,4,3,10,11,18,16,7],[12],[14],[6],[1],[2]\} \\
		d:\{[0,15,3,2,12,14,18,20,16,11,1,19,8,10,7,17],[6],[5],[13],[9],[4]\} \\
		e:\{[0,7,5,3,9,13,12,14,19,10,2,1,4,16,8,17],[15],[18],[6],[11],[20]\} \\
	\end{split}
\end{equation}

The proposed model was validated on the above class incremental orders in terms of $C^{0:15}$, $C^{16:20}$ and $C^{0:20}$ (all), respectively. Experimental results are shown in Table~\ref{table-ClassOrders}.
\begin{table}[htbp]
	\centering
	\caption{Performance on various class incremental orders on VOC 15-1 task in terms of mIoU (\%).}
	\setlength{\tabcolsep}{4.0mm}{
		\begin{tabular}{c|cc|c}
			\toprule[0.5mm]
			order&$C^{0:15}$&$C^{16:20}$&all\\
			\midrule
			a&75.00&37.52&66.08\\
			b&72.82&36.36&64.14\\
			c&70.73&37.38&62.79\\
			d&69.02&32.14&60.24\\
			e&74.05&46.46&67.48\\
			\midrule
			avg.$\pm$std.&-&-&64.15$\pm$2.53\\
			\bottomrule[0.5mm]
	\end{tabular}}
	\label{table-ClassOrders}
\end{table}
\begin{figure}[htbp]
	\centering
	\includegraphics[scale=0.42]{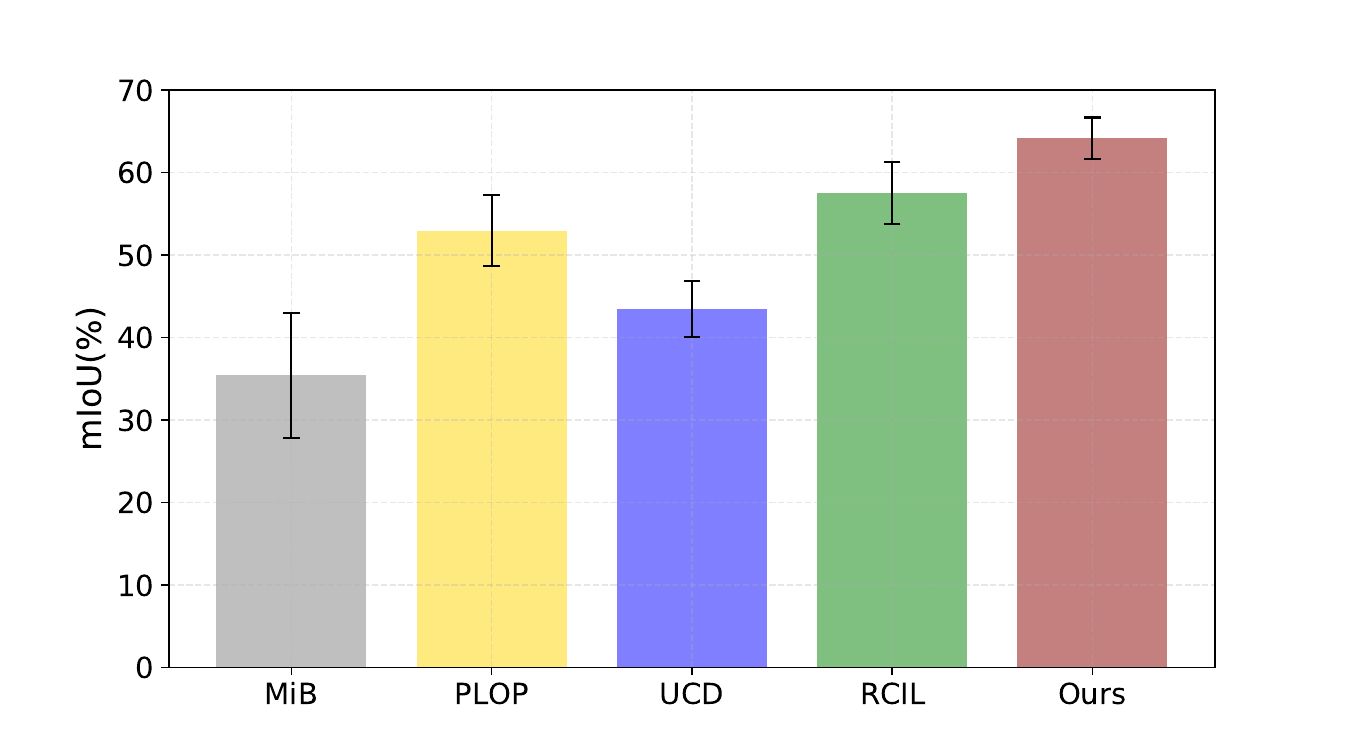}
	\caption{The average performance and standard variance under different incremental class orders on VOC 15-1 task.}
	\label{fig-mIoU_class_orders}
\end{figure}
We compare the average mIoU and standard variance on VOC 15-1. The experimental results in Fig.~\ref{fig-mIoU_class_orders} prove that our method achieves the highest average mIoU with the least standard deviation, surpassing four representative pioneers~\cite{MiB, PLOP, UCD, RCIL} in terms of robustness and quantitative performance.

\subsubsection{Failure Example and Analysis}
\label{Sec-Abla-Failure}
\begin{figure}[h]
	\centering
	\includegraphics[scale=0.55]{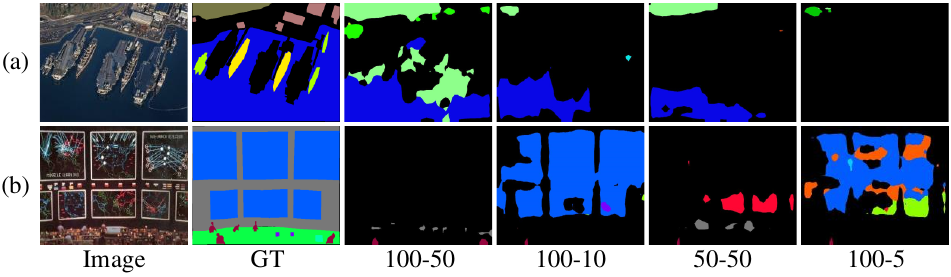}
	\caption{Failure examples of the proposed method.}
	\label{fig-failure_example}
\end{figure}

Above experiments validate the efficiency of the proposed model in three benchmarks. However, we found our model has limited performance in such conditions: 1) Long-tail incremental class: Since the small amount of data especially under data-limited conditions,  the model encounters difficulties in learning incremental classes and lack of robustness. For example in Fig.~\ref{fig-failure_example}(a), the ship is a long-tail class in ADE20K with infrequent viewing angles, leading to model degradation especially in multi-step tasks like 100-5 (11 steps). 2) Multi-class incremental learning: It is challenging to achieve multiple new classes learning at one incremental step, i.e., the semantic drift causes severe model degradation. Fig.~\ref{fig-failure_example}(b) shows a typical complex indoor scene. Our model fails to perform accurate pixel-level segmentation especially in multi-class incremental learning like ADE 100-50 and ADE 50-50 tasks.  

\section{Conclusion and Discussion}
This paper addresses continual semantic segmentation from an efficient, robust, human-like and interpretable perspective. This paper mainly presents three contributions, 1) A disentangled distillation method is proposed, the knowledge is decoupled to semantic-invariant and sample-specific terms with specified loss constraint for solid knowledge inheritance. 2) A generic interpretability method is proposed for CSS tasks, which is suitable and advantageous to reveal the knowledge transfer and model updating effect. 3) An end-to-end architecture is proposed, which is model-agnostic and without exemplar-memory. With all-sided experiments, validation and analysis, the proposed method achieves solid performance in multiple datasets. In particular, the proposed model has superior efficient in data-limited conditions and in challenging CSS tasks, such as class\&domain incremental scenarios and multi-step CSS settings.

However, the limitations of the proposed model mainly lie in prototypes confusion, i.e., the class-wise prototypes may face semantic confusion in semantic enrichment scenarios among similar classes. And it brings a little extra storage consumption for prototypes. In addition, the interpretability analysis and modelling can be continually investigated. Our future work will continue to explore the coupling between human cognition patterns and machine learning models, and the few-shot/zero-shot incremental learning in open-world scenarios.

\ifCLASSOPTIONcompsoc
  \section*{Acknowledgments}
\else
  \section*{Acknowledgment}
\fi
This work was supported by the National Natural Science Foundation of China under Grant 62271018.

\ifCLASSOPTIONcaptionsoff
  \newpage
\fi

{\small
	\bibliographystyle{IEEEtran}
	\bibliography{refs}
}

\end{document}